\documentclass{article}

\usepackage[preprint]{neurips_2026}

\usepackage[utf8]{inputenc} 
\usepackage[T1]{fontenc}    
\usepackage{hyperref}       
\usepackage{url}            
\usepackage{booktabs}       
\usepackage{amsfonts}       
\usepackage{nicefrac}       
\usepackage{microtype}      
\usepackage{xcolor}         

\usepackage{amsmath}
\usepackage{amssymb}
\usepackage{mathtools}
\usepackage{amsthm}
\usepackage{graphicx}
\usepackage{subcaption}
\usepackage{multirow}

\title{Training a Predictive Coding Network on ImageNet \\ using Equilibrium Propagation}

\author{
  Tugdual Kerjan\thanks{Equal contribution.} \\
  Rain AI \\
  \texttt{tkerjan@outlook.com} \\
  \And
  Rasmus Høier\footnotemark[1] \\
  Rain AI \\
  \texttt{rasmus@rain.ai} \\
  \AND
  Benjamin Scellier\footnotemark[1] \\
  Rain AI \\
  \texttt{benjamin@rain.ai} \\
}

\begin{document}

\maketitle

\begin{abstract}
Equilibrium Propagation (EP) is a physics-based training framework that has primarily been employed in
energy-based models, including continuous Hopfield networks, nonlinear resistive networks and coupled phase oscillators. However, EP's practical applications have so far remained limited to relatively small-scale problems. Predictive coding networks (PCNs), another class of energy-based models rooted in computational neuroscience, are typically trained with a specialized algorithm and have likewise not yet been demonstrated at large scale. In this work, we develop an EP-based training method for PCNs which combines the centered variant of EP with a novel equilibration scheme for PCNs. Using this approach, we train a 10-layer convolutional PCN (VGG10) on full-size ImageNet, achieving 13.23\% test error rate on the top-5 classification task, close to the 12.2\% backpropagation baseline. To our knowledge, this is the first demonstration of both PCNs and EP-based training at ImageNet scale. These results significantly extend the scalability of both approaches and suggest that the primary challenges in scaling EP in other physical systems may come more from the computational properties of these systems than from inherent limitations of the EP framework.
\end{abstract}

\section{Introduction}

The growing computational demands of machine learning (ML) motivate the exploration of neuromorphic computing platforms based on analog physics, which promise substantial energy savings \citep{markovic2020physics}. A prominent research direction within these efforts is the development of `physical learning algorithms' that leverage the physics of the hardware for both inference and learning \citep{momeni2025training}. Equilibrium Propagation (EP) \citep{scellier2017equilibrium} is a physics-based training framework that enables a class of systems to learn by adjusting parameters through comparisons of equilibrium states under varying boundary conditions. 

EP belongs to a broader class of contrastive learning algorithms \citep{hinton1983optimal, movellan1991contrastive, baldi1991contrastive, hinton2002training} and employs the concept of weak perturbations used in earlier works \citep{hinton1987learning,o1996biologically}. EP distinguishes itself from these methods in that it allows optimization of arbitrary cost functions, and in the weak-perturbation regime it approximates gradient descent. EP has been studied in energy-based models such as continuous Hopfield networks \citep{scellier2017equilibrium}, nonlinear resistor networks \citep{kendall2020training}, flow and elastic networks \citep{stern2021supervised}, and systems of Kuramoto oscillators \citep{zoppo2022equilibrium,wang2024training,rageau2025training}. Generally applicable to physical systems governed by variational principles, EP has been extended to stochastic and thermal systems \citep{scellier2017equilibrium,massar2024equilibrium}, Lagrangian dynamical systems \citep{kendall2021gradient,scellier2021deep,massar2025equilibrium,pourcel2025lagrangian} and quantum systems \citep{massar2024equilibrium,wanjura2025quantum,scellier2024quantum}. Further developments include a version of EP which performs parameter updates in black-box systems using physical dynamics \citep{scellier2022agnostic} and a version of EP for nonlinear wave systems \citep{sajnok2025near}.

Despite these theoretical developments, applications of EP have so far remained limited to small networks and datasets. Physical implementations of EP and variants have been demonstrated on resistor networks \citep{dillavou2022demonstration,dillavou2024machine}, memristor networks \citep{yi2023activity}, D-wave's Ising machine \citep{laydevant2024training}, and elastic networks \citep{altman2024experimental}, albeit only at very small scales. On the simulation side, EP has been scaled to train convolutional Hopfield networks on a 32x32 downsampled version of ImageNet \citep{laborieux2022holomorphic,nest2024towards}. Setting aside the challenges of hardware implementation, several factors contribute to the difficulty of scaling EP in numerical experiments. First, unlike backpropagation, EP computes only an approximate gradient of the cost function, with its accuracy depending on factors like perturbation strength, finite difference scheme and the proximity to equilibrium. Second, the expressivity and computational properties of EP-compatible physical networks remain poorly understood, as these systems lack a direct mapping to traditional neural networks \footnote{\citet{scellier2025universal} recently established a universal approximation result for nonlinear resistive networks.}. Finally, simulations of such networks can be computationally expensive, since they require numerical optimization of the system's energy function to reach equilibrium\footnote{Algorithmic advances have accelerated this equilibration step in some systems \citep{scellier2023energy,scellier2024fast}.}. As one attempts to scale EP to larger networks and more complex datasets, it has remained unclear whether the observed challenges should be attributed to the EP framework itself or the expressivity of the underlying networks (or a combination of both).

In this work, we investigate this question by applying EP to predictive coding networks (PCNs), a class of models studied in computational neuroscience that alleviates some of the challenges associated with simulations of physical networks. In their most common form, PCNs can be viewed as an energy-based counterpart to feedforward neural networks~\citep{whittington2017approximation, millidge2021predictive}. Although it is unclear whether PCNs can offer advantages as a neuromorphic platform \citep{zahid2023predictive}, in numerical experiments they allow to isolate EP-related challenges from model-specific issues: at test time, PCNs behave like feedforward neural networks, enabling direct comparison with backpropagation. PCNs also offer a computational advantage over simulations of physical systems such as resistor or oscillatory networks, as their `free equilibrium' state (i.e. the inference equilibrium) can be obtained in a single forward pass.

To date, however, applications of PCNs have also been limited to relatively small datasets such as Tiny ImageNet \citep{pinchetti2025benchmarking}, and although some works on PCNs have explored the nudging-based perturbation mechanism of EP \citep{whittington2019theories,millidge2023backpropagation}, PCNs are more typically trained using a different perturbation strategy based on output clamping. To identify the key ingredients for effectively training PCNs with EP, we conduct extensive experiments on a 5-layer convolutional PCN (VGG5) across four vision datasets. We perform a comprehensive sensitivity analysis of EP hyperparameters, including the perturbation method, finite difference scheme, perturbation strength and the number of iterations in the nudge-phase equilibration. Our main findings and contributions are as follows:

\begin{itemize}
\item While clamping-based perturbations perform comparably to EP’s nudging-based approach on CIFAR-100, we find that the nudging-based method significantly outperforms clamping on more challenging datasets such as ImageNet 32x32 (Figure~\ref{fig:perturbation_methods}).
\item By combining the nudging-based approach with the centered scheme, we train a 10-layer convolutional PCN (VGG10) on full-resolution ImageNet, achieving 13.23\% top-5 test error rate, close to the 12.2\% backpropagation baseline (Table~\ref{table:vgg10-results}). To the best of our knowledge, this is the first demonstration of both PCNs and EP-based training at ImageNet scale.
\item Our results challenge common assumptions about EP's random and centered schemes, showing that the random scheme remains competitive even on full-size ImageNet (Figure~\ref{fig:perturbation_methods} and Table~\ref{table:vgg10-results}). This scheme is particularly appealing for PCNs, as the learning rule involves only a single equilibrium state, rather than two as in the centered scheme.
\end{itemize}


\section{Related Work on Predictive Coding Networks}

The standard learning algorithm for PCNs as outlined in \citet{whittington2017approximation}, involves clamping the output units to desired outputs rather than nudging them as in EP. \citet{millidge2023backpropagation} proposed using the nudging method of EP in the context of PCNs, but they used specifically the forward scheme (which employs positive nudging) and found the standard learning algorithm for PCNs to be preferable. Conversely, \citet{pinchetti2025benchmarking} combined the weak clamping-based perturbation approach with the random scheme and achieved significantly better performance, using it to train convolutional PCNs on Tiny ImageNet. The centered scheme for EP \citep{laborieux2021scaling}, which involves comparing one positively-perturbed state with one negatively-perturbed state, has not been utilized in PCNs so far, presumably because it requires comparing two network states, making it harder to justify from a neuroscience perspective. Our paper is the first to explore EP's nudging approach combined with the backward, random and centered schemes in the context of PCNs.

Scaling PCNs to deeper architectures has been a key challenge~\citep{pinchetti2025benchmarking}, with a number of recent papers exploring this issue. \citet{innocenti2026mu} improved activation stability by introducing additional scaling factors in the network energy, enabling the training of densely connected ResNets with up to 128 layers on MNIST and Fashion-MNIST. Other works have focused on tackling the issue of signal attenuation due to limited machine precision. \citet{goemaere2025error} tackle this issue by reparametrizing PC to treat the local errors as the variable of optimization rather than the neural states, showing that the underlying optimization problems are equivalent, and the simulations are much faster (more than 100$\times$). \citet{qi2025towards} address the issue of signal attenuation by, among other things, introducing a schedule for the inference step size and using a surrogate objective when computing the weight updates.

\section{Background: Equilibrium Propagation}
\label{sec:equilibrium-propagation}

This section reviews EP and its variants depending on finite difference schemes (forward, backward, centered and random) and perturbation methods (nudging vs clamping).

In its original formulation, EP applies to \emph{energy-based systems}, meaning systems with a state vector $h$ whose dynamics evolve towards a minimum (or more generally critical point) of some function $E$, referred to as \emph{energy function}. In supervised learning, $E(\theta,x,h)$ depends on trainable weights $\theta$ and a boundary input $x$. During inference, given $x$, the system settles into its equilibrium state $h(\theta,x)$, characterized by the first-order condition
\begin{equation}
\label{eq:equilibrium-state}
\frac{\partial E}{\partial h} (\theta,x,h(\theta,x)) = 0.
\end{equation}
The function $x \mapsto h(\theta,x)$ represents the system’s forward computation. Training consists in adjusting $\theta$ so as to minimize the discrepancy between the prediction $h(\theta,x)$ and the desired result $y$, which is quantified using a cost function $C(h(\theta,x),y)$. The training objective we seek to minimize is
\begin{equation}
\label{eq:objective}
J(\theta) = \mathbb{E}_{(x,y)} \left[ C(h(\theta,x),y) \right],
\end{equation}
where the expectation is taken over the data distribution of input-target pairs $(x,y)$.
If, given an input-output pair $(x,y)$ from the training data, one had direct access to $\nabla_\theta C(h(\theta,x),y)$, the usual gradient-descent update would be
\begin{equation}
\label{eq:gradient-descent}
\Delta \theta = - \eta \nabla_\theta C(h(\theta,x),y).
\end{equation}
The key challenge in a physical system is to estimate these gradients using the system's dynamics itself. EP accomplishes this by introducing a small \textit{nudging} parameter $\beta \in \mathbb{R}$, and interpreting $\beta C(h,y)$ as an interaction energy between the state $h$ and target output $y$, scaled by $\beta$. Incorporating this into the energy function leads to the \textit{total energy function}
\begin{equation}
\label{eq:total-energy-function}
F(\theta,\beta,h) = E(\theta,x,h) + \beta C(h,y),
\end{equation}
where $\beta$ controls the strength of this interaction. With $\beta=0$, the system relaxes to the \textit{free} equilibrium, $h_\star^0 = h(\theta,x)$. For nonzero $\beta$, the system settles to the \emph{nudged} equilibrium, characterized by the first-order condition
\begin{equation}
\label{eq:nudge-state}
\frac{\partial F}{\partial h}(\theta,\beta,h_\star^\beta) = 0.
\end{equation}
The main insight of \citet{scellier2017equilibrium} is that the gradient of the cost function with respect to the trainable weights can be approximated as
\begin{equation}
\label{eq:ep-formula}
\nabla_\theta C(h(\theta,x),y) = \left. \frac{d}{d\beta} \frac{\partial F}{\partial \theta}(\theta,\beta,h_\star^\beta) \right|_{\beta=0} \approx \frac{1}{\beta} \left[ \frac{\partial F}{\partial \theta}(\theta,\beta,h_\star^\beta) - \frac{\partial F}{\partial \theta}(\theta,0,h_\star^0) \right].
\end{equation}
Substituting Eq.~\eqref{eq:ep-formula} into Eq.~\eqref{eq:gradient-descent} yields a contrastive learning rule.

\subsection{Finite Difference Schemes: Forward, Backward, Centered and Random Schemes}

In contrast with standard ML methods based on automatic differentiation, EP does not provide the exact cost gradient but rather an approximation of it, due to the finite difference scheme employed for estimating the derivative with respect to $\beta$. We refer to the right-hand side of Eq.~\eqref{eq:ep-formula} with $\beta>0$ as the forward scheme. \citet{scellier2017equilibrium} found that the random scheme, where the sign of $\beta$ is chosen at random for each training pair $(x,y)$, works better than the forward scheme. To further reduce the approximation error, \citet{laborieux2021scaling} introduced the centered scheme whose error term is of order $O(\beta^2)$,
\begin{equation}
\label{eq:eqprop-centered}
\nabla_\theta C(h(\theta,x),y) \approx \frac{1}{2\beta} \bigg( \frac{\partial F}{\partial \theta} \left( \theta,\beta,h_\star^\beta \right) - \frac{\partial F}{\partial \theta} \left( \theta,-\beta,h_\star^{-\beta} \right) \bigg),
\end{equation}
and found it to yield better empirical results than the random scheme. The purely backward scheme, employing Eq.~\eqref{eq:ep-formula} with $\beta<0$, was first studied by \citet{scellier2022agnostic} and shown to optimize an upper bound of the objective function (Eq.~\eqref{eq:objective}).

\subsection{Perturbation Approaches: Nudging vs Clamping}

An alternative to the nudging approach of Eq.~\eqref{eq:total-energy-function} is to perturb the system by clamping its outputs rather than modifying its energy function. Let $h = (h_{\rm hid}, h_{\rm out})$ denote the system state, partitioned into hidden and output components. In the clamping-based approach, the output variables are set to
\begin{equation}
h_{\rm out}^\beta = (1-\beta) h_{\rm out}^0 + \beta y,
\end{equation}
where $h_{\rm out}^0$ denotes the free-equilibrium output of the system. This strategy, which predates EP, was proposed by \citet{o1996biologically} in the context of Hopfield-like networks, and has more recently been employed in physical energy-based networks \citep{stern2021supervised} and in PCN experiments \citep{pinchetti2025benchmarking}.

Unlike nudging, which provides flexibility in defining the learning objective of Eq.~\eqref{eq:objective} by allowing arbitrary choices of cost functions ($C$), the clamping-based approach does not define a cost function explicitly, and the resulting parameter updates do not generally correspond to gradient descent or to the minimization of an objective function, even in the limit $\beta \to 0$.
We refer to \citet{scellier2023energy} for a detailed comparison of the two approaches.

In numerical experiments with convolutional Hopfield networks, \citet{scellier2023energy} found the nudging-based approach to outperform clamping. As we show in Section~\ref{sec:experiments}, the same conclusion holds for PCNs.

\section{Predictive Coding Networks Through the Lens of EP}
\label{sec:ff-energy-function}

While EP’s main appeal lies in training physical systems such as those referenced in the introduction, here we apply EP to feedforward neural networks by defining an appropriate energy function. This leads to a collection of algorithms for Predictive Coding Networks (PCNs).

Consider a feedforward neural network with $L$ layers:
\begin{equation}
\label{eq:feedforward-neural-network}
h^\star_0=x, \quad \text{and} \quad h^\star_k = f_k(\theta_k,h^\star_{k-1}), \quad 1 \leq k \leq L,
\end{equation}
where $h_k^\star$ is the activation of layer $k$, $f_k$ is the (learnable) transformation that maps $h_{k-1}^\star$ to $h_k^\star$, and $\theta_k$ are the corresponding trainable parameters (weights and biases). Later we will consider specifically convolutional (VGG) networks, but for now our discussion does not make such assumptions on the network architecture. One can view this network as an energy-based model whose energy function is
\begin{equation}
\label{eq:ff-compatible-energy-function}
E_{\rm PCN}(\theta,x,h) = \frac{1}{2} \sum_{k=1}^L \left\| h_k - f_k(\theta_k,h_{k-1}) \right\|^2,
\end{equation}
where $x=h_0$ and $h=(h_1,\ldots,h_L)$. Indeed, the state $h^\star=(h_1^\star,h_2^\star,\ldots, h_N^\star)$ satisfying the feedforward equations of Eq.~\eqref{eq:feedforward-neural-network} is a stationary point of $E_{\rm PCN}$, specifically the unique global minimum of $E_{\rm PCN}$, for which we have $E_{\rm PCN}(\theta,x,h^\star)=0$.

Assuming we use the nudging-based perturbation technique, EP then operates on the total energy $F_{\rm PCN}(\theta,\beta,h) = E_{\rm PCN}(\theta,x,h) + \beta C(h_L,y)$, where $C(h_L,y)$ is the cost function. In the \emph{free phase} ($\beta=0$), finding the global minimum of $F_{\rm PCN}$ with respect to $h$ reduces to a single forward pass, following the feedforward equations of Eq.~\eqref{eq:feedforward-neural-network}. For $\beta \neq 0$ (nudge phase), however, finding a stationary point of $F_{\rm PCN}$ requires a specific algorithm. In this work, we use a modified form of gradient descent on $F_{\rm PCN}$ with respect to $h$, which we describe next (Section ~\ref{sec:nudge-phase-dynamics}). Besides, it is easy to verify that the energy derivative at the free equilibrium state vanishes: $\frac{\partial F_{\rm PCN}}{\partial \theta}(\theta,0,h_\star^0) = 0$. Therefore, when employing the forward, backward or random scheme, the EP gradient of Eq.~\eqref{eq:ep-formula} simplifies, requiring only one equilibrium state (the nudge state):
\begin{gather}
\nabla_\theta C(h(\theta,x),y) \approx \frac{1}{\beta} \frac{\partial F_{\rm PCN}}{\partial \theta}(\theta,\beta,h_\star^\beta).
\end{gather}

Prior works discussing the relationship between EP and PCNs combined the nudging approach with the forward scheme \citep{millidge2023backpropagation}, or the forward, backward and random schemes with the clamping approach \citep{pinchetti2025benchmarking}. Here, we perform a more comprehensive comparison of these perturbation techniques and finite difference schemes. We find that the best performing ones combine the nudging approach with the centered, random and backward schemes, which were not used in prior works on PCNs. Appendix~\ref{sec:traditional-pcn-training} provides more details on the relationship between EP and the traditional learning algorithm for PCNs.

\subsection{Nudged-Phase Equilibration via Projected Gradient Descent}
\label{sec:nudge-phase-dynamics}

To compute the nudge equilibria, we minimize the total energy function using projected gradient descent (PGD). The PCN's total energy function rewrites
\begin{equation}
F_{\rm PCN}(\theta,\beta,h) = \frac{1}{2} \sum_{k=1}^L \| \epsilon_k \|^2 + \beta C(h_L,y), \qquad \epsilon_k = h_k - f_k(\theta_k, h_{k-1}),
\end{equation}
where $\epsilon_k$, referred to as \textit{error}, is the difference between the state $h_k$ and the bottom-up prediction $f_k(\theta_k,h_{k-1})$. In the nudge-phase dynamics, units $h_k$ adjusts so as to minimize the sum of squared errors. In what follows, we consider PCNs with ReLU units in the hidden layers and linear output units, more specifically:
\begin{align}
f_k & \in\{ \text{ReLU}\circ \text{maxpool} \circ \text{conv2d}, \; \text{ReLU}\circ \text{conv2d}, \; \text{ReLU}\circ \text{matmul}\}, \qquad 1 \leq k \leq L-1, \\
f_L &= {\rm matmul}.
\end{align}
These are the operations we require for the VGG networks considered in our experiments.

PGD with step size $\alpha>0$ on a ReLU layer reads
\begin{equation}
\label{eq:pgd-step}
h_k \leftarrow {\rm Proj}_{[0,+\infty]} \left(h_k - \alpha  \frac{\partial F_{\rm VGG}}{\partial h_k}(\theta, \beta, h) \right),
\end{equation}
where ${\rm Proj}_{[0,+\infty]}$ is the projection onto the range of the ReLU, that is $[0,+\infty]$, which is equivalent to applying the ReLU itself. In our experiments, we use step size $\alpha=1$ (see Appendix~\ref{sec:nudge-phase-equilibration} for a justification). For a hidden layer $h_k$, the gradient of $F_{\rm VGG}$ with respect to $h_k$ is
\begin{equation}
\label{eq:aux}
\frac{\partial F_\text{VGG}}{\partial h_k} = \frac{\partial \epsilon_k^2}{\partial h_k} + \frac{\partial \epsilon_{k+1}^2}{\partial h_k} = h_k - f_k(\theta_k, h_{k-1}) + \frac{\partial \epsilon_{k+1}^2}{\partial h_k}.
\end{equation}
Injecting Equation~\eqref{eq:aux} into Equation~\eqref{eq:pgd-step} with $\alpha=1$, the $h_k$ terms cancel out and the PGD update rewrites more concisely as
\begin{equation}
h_k \gets \text{ReLU}\left( f_k(\theta_k, h_{k-1})
+ \frac{\partial \epsilon_{k+1}^2}{\partial h_k}\right).
\label{eq:pgd}
\end{equation}
For the output layer whose activation function is the identity, the update rule simplifies as:
\begin{equation}
h_L \gets a_L - \beta \frac{\partial C}{\partial h_L}(h_L,y).
\label{eq:output-units}
\end{equation}
Thus, during the nudged dynamics ($\beta \neq 0$), output units are perturbed by bringing them to either lower cost values ($\beta>0$) or higher cost values ($\beta<0$), that is, closer to the targets or further away from the targets. These perturbations lead to non-zero error signal $\epsilon_L$, which leads to perturbations in the layer $L-1$, and so on.

In practice, we find that a modified version of PGD works better than PGD itself -- see Appendix~\ref{sec:nudge-phase-equilibration}.

We emphasize that it remains unclear whether $F_{\rm PCN}$ and the nudge-phase dynamics (PGD) could be efficiently realized in hardware, to obtain the nudge equilibrium using physical dynamics. The primary contribution of our work is to advance the state of EP-based and PCN numerical experiments (on GPUs) to full-size ImageNet, without hardware implementation considerations.

\section{Experimental Results}
\label{sec:experiments}

We consider two convolutional (VGG) PCN models: a 5-layer network (VGG5) with 32x32 pixel input images, and a 10-layer network (VGG10) with 224x224 pixel input images. The VGG5 network is employed on MNIST, CIFAR10, CIFAR100 and ImageNet 32x32 (the 32x32 pixel downscaled version of ImageNet). To permit using VGG5 on MNIST, images from MNIST are padded to 32x32 input size. The VGG10 network is trained on the (full-size) ImageNet dataset. Network architectures and numerical simulation details are described in Appendix~\ref{sec:simulation-details}.

The hyperparameter search space for EP-based training of these VGG networks includes:
\begin{itemize}
\item perturbation method: nudging vs clamping,
\item finite difference scheme: forward, backward, centered or random,
\item nudging strength ($\beta$),
\item number of nudge iterations ($K$),
\item cost function (only applicable to the nudging-based perturbation approach): mean squared error (MSE) or cross-entropy (CE).
\end{itemize}
Backpropagation (BP) is also employed as a baseline for comparison. We use a two-stage hyperparameter tuning strategy, where we first tune the hyperparameters shared by BP-based and EP-based training (weight initialization schemes, learning rates, weight decay, batch-size, cost function, etc), and subsequently tune the EP-specific hyperparameters. This strategy is justified by the fact that, when employing the nudging-based perturbation method, in the limit when $\beta$ tends to $0$, the EP gradients are equal to the BP gradients (Equation~\eqref{eq:ep-formula}).

We find that the best performing version of EP involves the nudging-based perturbation approach and the centered scheme. Unless stated otherwise, and `EP-trained network' refers to this configuration.

\subsection{Comparison of Perturbation Approaches and Finite Difference Schemes}

\begin{figure}[h]
    \centering
    \begin{subfigure}[b]{0.495\linewidth}
        \centering
        \includegraphics[width=\linewidth]{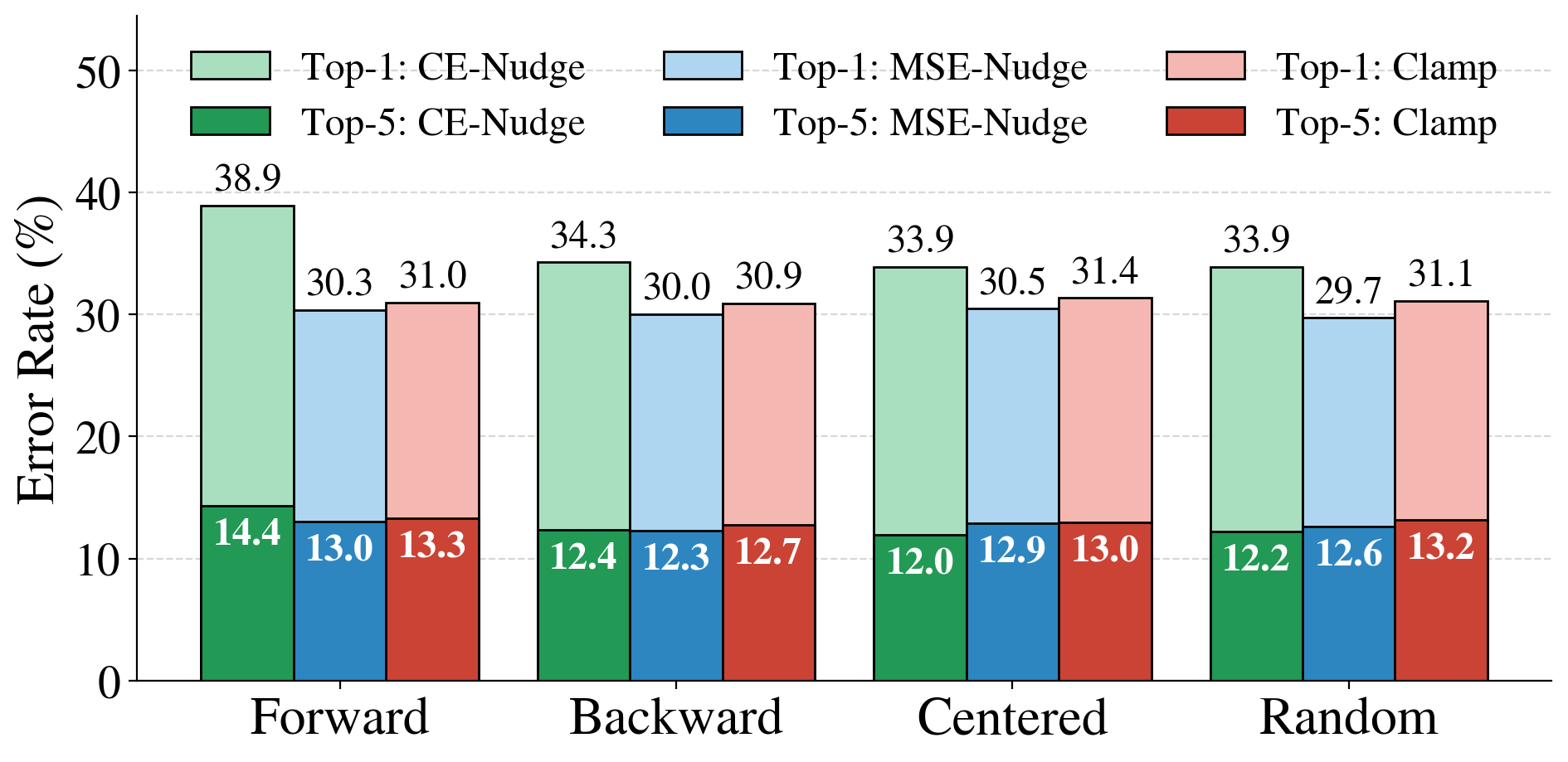}
        \caption{CIFAR-100}
        \label{fig:perturbation_cifar100}
    \end{subfigure}
    \hfill
    \begin{subfigure}[b]{0.495\linewidth}
        \centering
        \includegraphics[width=\linewidth]{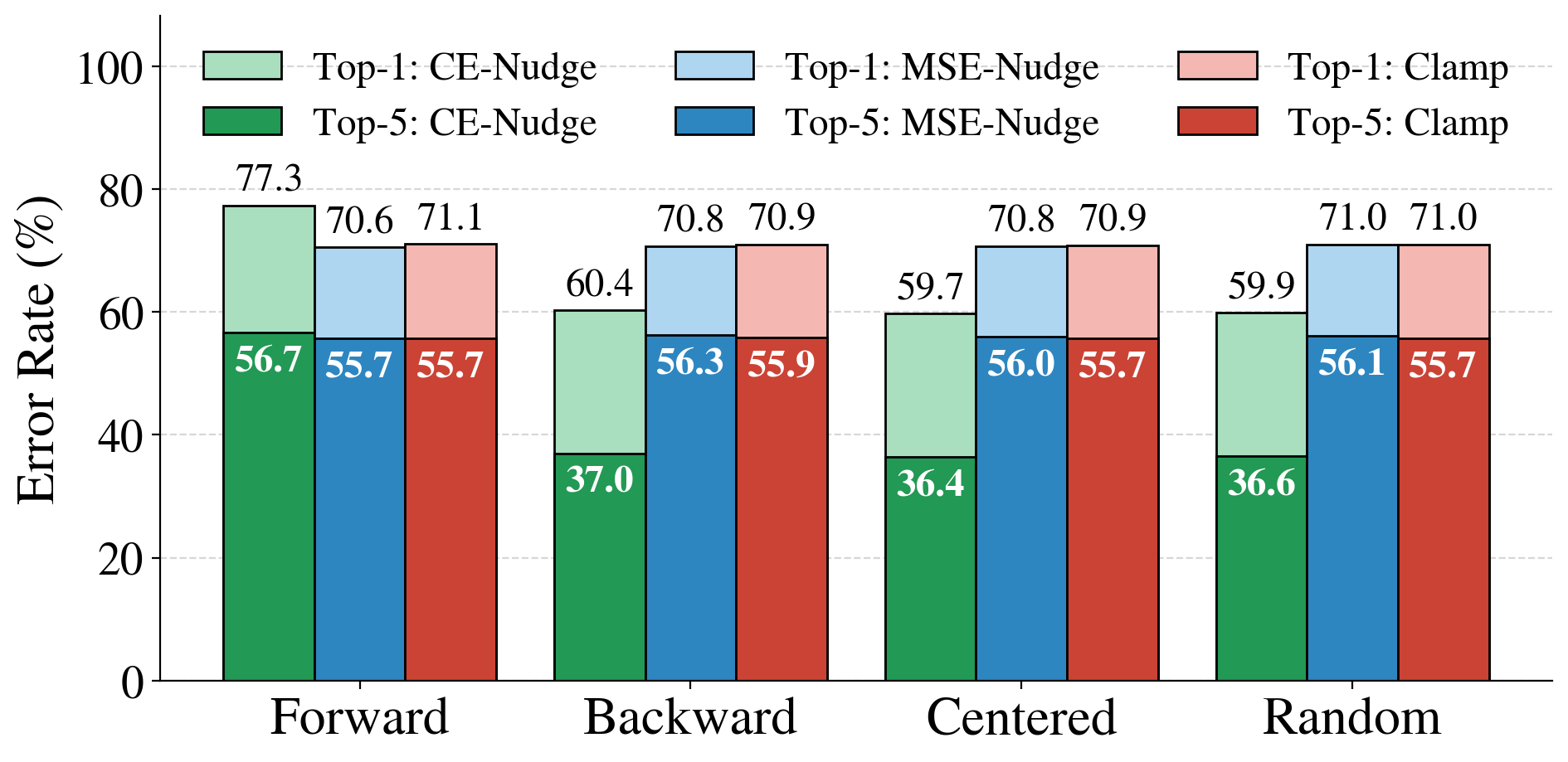}
        \caption{ImageNet 32x32}
        \label{fig:perturbation_imagenet32}
    \end{subfigure}
    \caption{Comparison of perturbation approaches (CE-Nudge, MSE-Nudge and Clamp) and finite difference schemes (forward, backward, centered and random) on the VGG5 network trained on (a) CIFAR-100 and (b) ImageNet 32x32.}
    \label{fig:perturbation_methods}
\end{figure}

We compare the two perturbation approaches (nudging and clamping) and the four finite difference schemes (forward, backward, centered and random). For nudging-based perturbation, we use both the MSE and cross-entropy (CE) cost functions.

Figure~\ref{fig:perturbation_methods} reports the results for the VGG5 trained on CIFAR100 and ImageNet 32x32. While all algorithms perform comparably on CIFAR100, important differences arise on the more challenging ImageNet 32x32 task. CE-Nudge performs significantly better than both MSE-Nudge and Clamp, except when employed in combination with the forward scheme. The best performing method is CE-Nudge with the centered scheme, in agreement with prior literature on EP \citep{laborieux2021scaling,scellier2023energy}, but interestingly, both the random and backward schemes perform comparably with the centered scheme, which challenges the conclusions of these works, where more important gaps were observed.

\subsection{Sensitivity to Nudging Strength and Number of Nudge Iterations}

\begin{figure}[htbp]
    \centering
    \begin{minipage}[t]{0.56\textwidth}
        \vspace{0pt} 
        \centering
        \includegraphics[width=\linewidth]{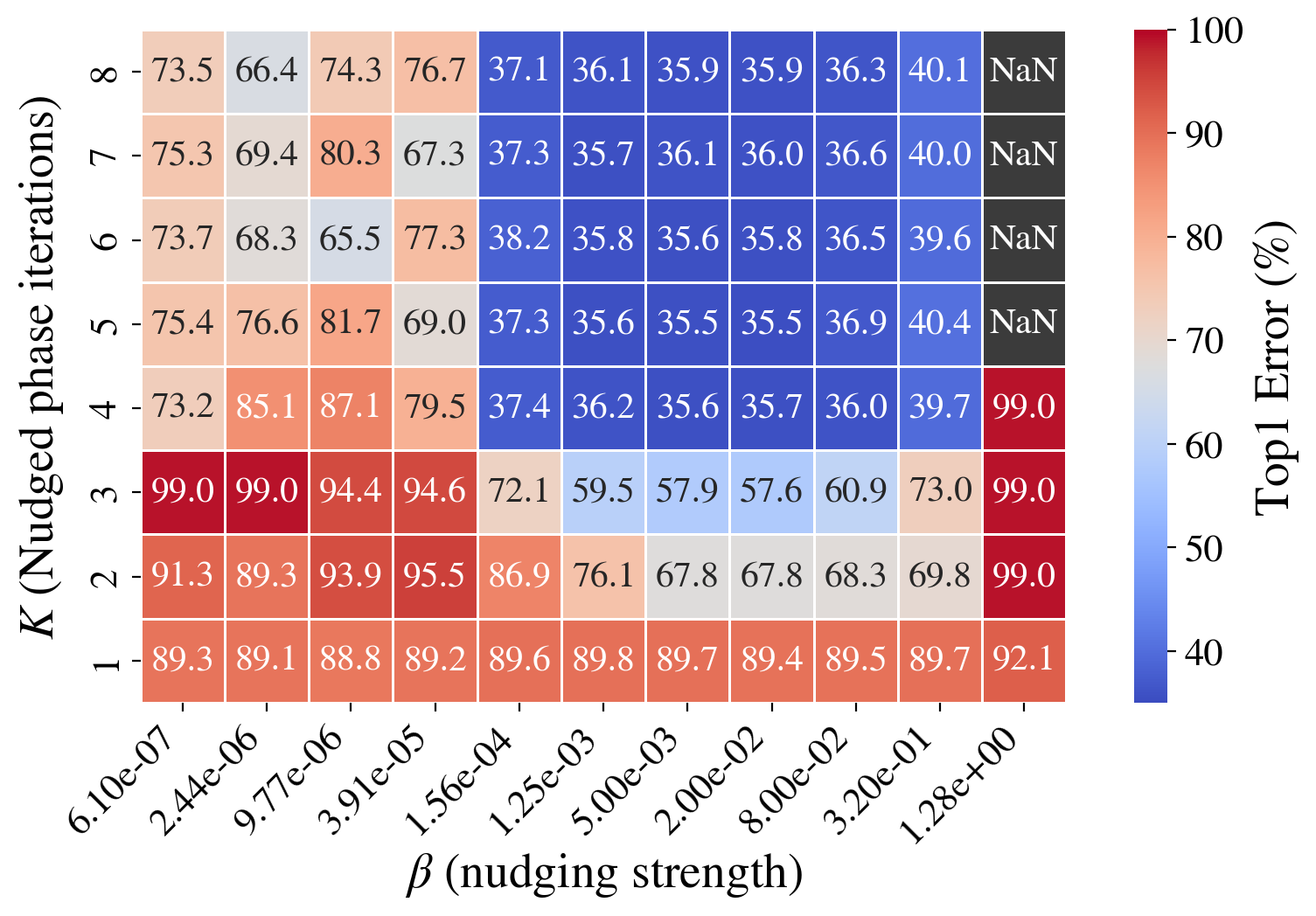}
        \caption{VGG5 network trained on CIFAR100 for 20 epochs. Sensitivity to the nudging strength ($\beta$) and number of nudge iterations ($K$).}
        \label{fig:sensitivity}
    \end{minipage}\hfill
    \begin{minipage}[t]{0.42\textwidth}
        \vspace{0pt} 
        \centering
        \includegraphics[width=\linewidth]{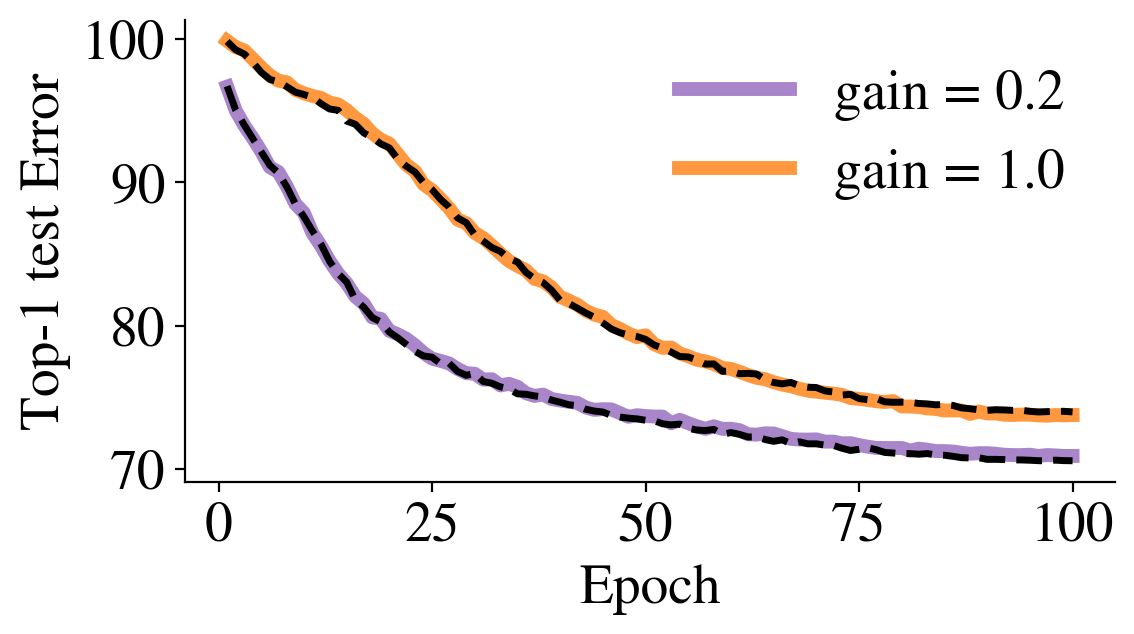}
        \caption{VGG5 trained on ImageNet 32x32 using the MSE cost function. Different weight initialization gains for the output layer lead to significantly different results ; however, EP consistently performs comparably to BP. Solid lines and dashed lines denote EP  and BP, respectively.}
        \label{fig:vgg5-weight-init}
    \end{minipage}
\end{figure}

Figure~\ref{fig:sensitivity} displays the performance of a trained VGG5 network for different values of $K$ (number of nudged iterations) and $\beta$ (nudging strength). We observe that, as long as $K \geq 4$ and $0.0002 \leq \beta \leq 0.1$, the final error rate is mostly insensitive to the choice of $K$ and $\beta$.

Intuitively, $K$ needs to be large enough that the network reaches equilibrium, but should be chosen as small as possible to avoid unnecessary computation. Empirically, the optimal value for $K$ is the number of layers in the network, allowing perturbations to fully propagate from outputs to inputs. Similarly, $\beta$ needs to be small enough that EP gradients are accurate (Equation~\eqref{eq:ep-formula}), but it needs to be large enough that perturbations can propagate throughout the network without vanishing (due to limited machine precision).

Appendix~\ref{sec:nudge-phase-equilibration} provides more insights into the nudge-phase equilibration dynamics, including sensitivity to the step size ($\alpha$) and the traversal scheme, as well as equilibration curves ($F$ as a function $K$).

\subsection{Sensitivity to Batch Size, Cost Function and Weight Initialization}

\begin{figure}[h]
\centering
\begin{subfigure}[b]{0.49\textwidth}
    \centering
    \includegraphics[width=\textwidth]{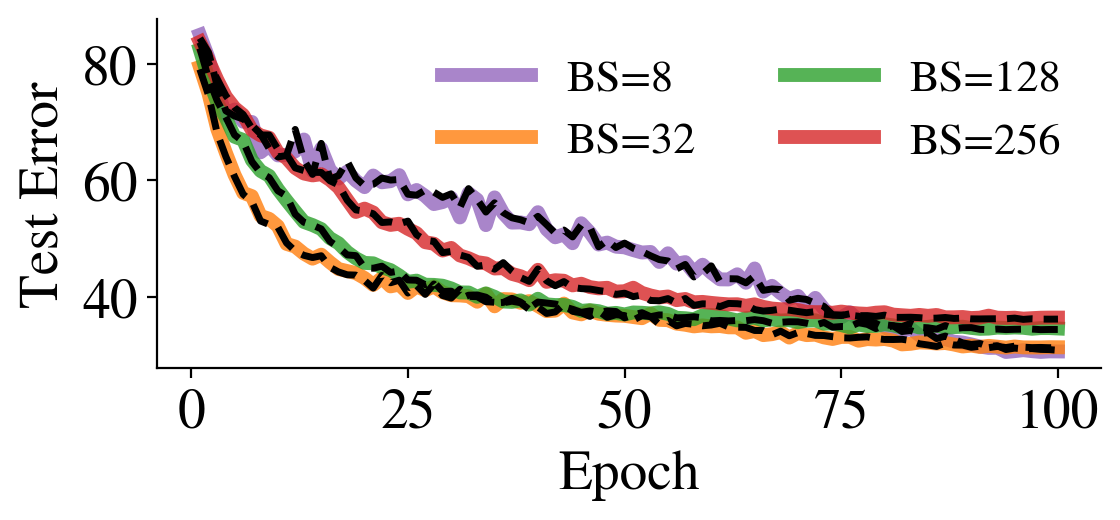}
    \caption{Test error rate}
    \label{fig:vgg5-batch-size-test}
\end{subfigure}
\hfill 
\begin{subfigure}[b]{0.49\textwidth}
\centering
    \includegraphics[width=\textwidth]{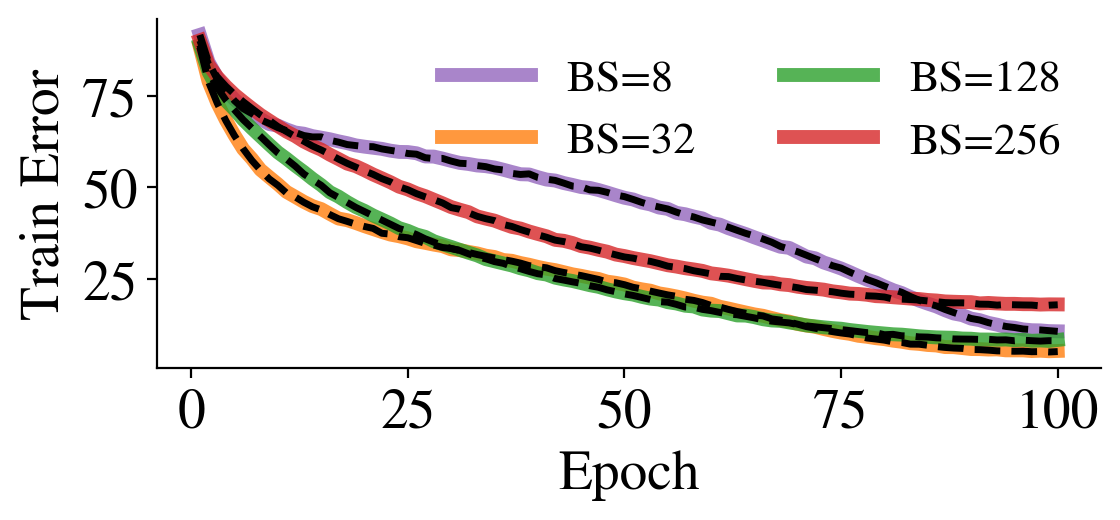}
    \caption{Train error rate}
    \label{fig:vgg5-batch-size-train}
\end{subfigure}
\caption{
VGG5 trained on CIFAR100 using the MSE. Shapes of training curves depend on the batch size (BS), but not on the algorithm (EP vs BP). Solid lines denote EP and dashed lines denote BP.
}
\label{fig:vgg5-batch-size}
\end{figure}

We now compare the performance of EP-trained PCNs against BP, and study their sensitivity to shared hyperparameters. Figure~\ref{fig:vgg5-batch-size} compares EP and BP on CIFAR-100 for different batch sizes, while Table~\ref{table:vgg5-results} reports the results of a more comprehensive comparison on four datasets (MNIST, CIFAR-10, CIFAR-100 and ImageNet 32x32) where we used two cost functions: the mean-squared error (MSE) and the cross-entropy (CE).

In every situation, EP is competitive with BP. While the end performance depends prominently on the cost function and the batch size, it depends little on the learning algorithm used (BP or EP).

For ImageNet 32x32 experiments with the MSE, we find (Figure~\ref{fig:vgg5-weight-init}) that significantly better results are obtained using a different weight initialization scheme (scaling the output layer's weights by 0.2 instead of the default 1.0). Here again, however, EP performs comparably with BP.

\begin{table}[ht]
\caption{Sensitivity to the choice of the cost function: mean squared error (MSE) and cross-entropy (CE). Top1 (resp. Top5) refers to the test error rate in \% for the classification task. In all cases, EP performs competitively with BP. Results were averaged over 3 random seeds.}
\label{table:vgg5-results}
\centering
\resizebox{\textwidth}{!}{
\begin{tabular}{cccccccc}
\toprule
\multirow{2}{*}{Cost} & \multirow{2}{*}{Algorithm} & \multirow{2}{*}{MNIST} & \multirow{2}{*}{CIFAR-10} & \multicolumn{2}{c}{CIFAR-100} & \multicolumn{2}{c}{ImageNet 32x32} \\ \cmidrule(r){5-6} \cmidrule(r){7-8}
& & & & Top1 & Top5 & Top1 & Top5 \\
\cmidrule(r){1-8}
\multirow{2}{*}{MSE} & EP & $0.32 \pm 0.01$ & $7.8 \pm 0.1$ & $30.1 \pm 0.0$ & $12.8 \pm 0.1$ & $70.8 \pm 0.2$ & $55.9 \pm 0.3$ \\
& BP & $0.33 \pm 0.00$ & $8.0 \pm 0.0$ & $30.1 \pm 0.4$ & $12.7 \pm 0.5$ & $70.7 \pm 0.1$ & $56.0 \pm 0.2$ \\
\cmidrule(r){1-8}
\multirow{2}{*}{CE} & EP & $0.45 \pm 0.01$ & $9.7 \pm 0.2$ & $34.7 \pm 0.1$ & $12.1 \pm 0.2$ & $60.0 \pm 0.1$ & $36.6 \pm 0.4$ \\
& BP & $0.48 \pm 0.02$ & $9.5 \pm 0.0$ & $34.2 \pm 0.4$ & $12.3 \pm 0.2$ & $59.8 \pm 0.0$ & $36.5 \pm 0.0$\\
\bottomrule
\end{tabular}
}
\end{table}

\subsection{Full-Size Imagenet Experiments}

Table~\ref{table:vgg10-results} reports the results (Top1 and Top5 error rates) obtained with the VGG10 network trained on full-size ImageNet. We compare these results with existing literature on PCNs and EP. For these experiments, we used the cross-entropy (CE) cost function. While the EP-trained network is competitive in performance with the BP-trained network, we observe a small gap. The VGG10 networks were trained for 50 epochs.

Crucially, no prior work on PCNs or EP has been carried out on full-size ImageNet classification. The best results available for PCNs were performed by \citet{pinchetti2025benchmarking,qi2025towards} on Tiny Imagenet, a dataset with ~13x less images (100,000 instead of 1.28 million), with a resolution ~12x smaller (64x64 instead of 224x224) and with 5x fewer classes (200 instead of 1000) than ImageNet. Likewise, the best available results for EP were reported on ImageNet 32x32~\citep{nest2024towards}, whose images are 50x smaller than the full-resolution version (32x32 instead of 224x224).

\begin{table}[h!]
\caption{Performance of VGG10 on full-size ImageNet, compared to the results available in the related literature on PCNs and EP on smaller ImageNet variants (Tiny ImageNet and ImageNet 32x32). For each dataset, we write for comparison the number of images, size of images and number of classes in this order. Top1 (resp. Top5) refers to the test error rate in \% for the top-1 (resp. top-5) classification task. Our results provide the first benchmark on full-size ImageNet for both EP-based training and PCNs.
}
\label{table:vgg10-results}
\centering
\begin{tabular}{cccc}
\toprule
Dataset & Model & Top1 & Top5 \\
\cmidrule(r){1-4}
Tiny ImageNet & \citet{pinchetti2025benchmarking} & 58.85 & 33.75 \\
(100K, 64x64, 200) & \citet{qi2025towards} & 44.69 & 20.70 \\
\cmidrule(r){1-4}
\multirow{2}{*}{ImageNet 32x32} & \citet{laborieux2022holomorphic} & 63.5 & 39.2 \\
\multirow{2}{*}{(1.28M, 32x32, 1000)} & \citet{hoier2023dual} & 58.52 & 35.10 \\
& \citet{nest2024towards} & 54.0 & 30.0 \\
\cmidrule(r){1-4}
\multirow{2}{*}{ImageNet} & VGG10 + EP + random scheme & 34.73 & 14.02 \\
\multirow{2}{*}{(1.28M, 224x224, 1000)} & VGG10 + EP + centered scheme & 33.81 & 13.23 \\
& VGG10 + BP (baseline) & 32.35 & 12.20 \\
\bottomrule
\end{tabular}
\end{table}

\subsection{Integrating Skip Connections}

To demonstrate the broader applicability of the methods to ResNet-like models, we perform experiments on a variant of the VGG10 with skip connections. It has a similar architecture as the VGG10 network but includes strided 1x1 convolutional skip-layer connections from layer 2 to layer 5, as well as from layer 5 to layer 8. We train this VGG10Skip model on full-size ImageNet and obtain similar performance as the VGG10 model. Specifically, centered EP achieves 33.92\% Top1 and 13.32\% Top5 test error rates, compared to 32.40\% and 12.1\% for the BP baseline.

We expect that better results could be obtained using deep ResNets with batch normalization, at the expense of significantly longer nudge-phase times. This is left for future work to investigate.

\section{Conclusion}

Numerical experiments of Equilibrium Propagation (EP) have so far primarily focused on Hopfield networks, with evaluations restricted to the 32x32-pixel version of ImageNet. Likewise, Predictive Coding Network (PCN) experiments have remained limited to datasets of the size of Tiny ImageNet. Using EP's nudging-based perturbation method and the centered scheme, we unlocked the training of a 10-layer convolutional PCN (VGG10) on full size ImageNet, providing the first benchmark for both EP-based training and PCNs on this dataset. Our EP results (13.23\% top-5 test error rate) approach the backpropagation baseline (12.20\%) and beat the seminal AlexNet benchmark (15.3\%)~\citep{krizhevsky2012imagenet}.

At smaller scale (VGG5), our sensitivity analysis to hyperparameters revealed new insights about both EP and PCNs. First we showed that, by allowing to choose the cost function to optimize, the nudging-based perturbation method of EP presents significant advantages over the clamping-based technique typically used in PCNs. Our results show that, while little to no difference is observed between the methods on CIFAR100, the cross-entropy-nudging technique leads to a significant improvement over both the MSE-nudging and the clamping techniques on the more challenging ImageNet 32x32 dataset. Second, our results challenge common assumptions on EP, showing that on ImageNet 32x32, the backward and random schemes both perform competitively with the centered scheme typically employed by default in EP experiments. The random scheme remains competitive with the centered scheme on full-size ImageNet. The backward and random schemes are especially attractive in the context of PCNs where energy derivatives at the free equilibrium state are zero, resulting in a single-phase algorithm, which also roughly halves the duration of experiments. Third, our results show that EP works even in the regime where the nudging parameter $\beta$ is tiny (up to $\beta \approx 0.0002$ in our experiments), much smaller than typical values used in earlier EP experiments.

Our extensive study conducted on the VGG5 model also revealed that, while classification accuracy varies depending on hyperparameters such as the cost function (mean squared error or cross-entropy), weight initialization and the batch size, in every situation the performance obtained with EP is comparable to BP. These results position EP as a viable alternative to BP for SGD-based optimization (stochastic gradient descent), while also highlighting that SGD is only one component of an effective learning system, with other factors - such as the choice of cost function, initialization scheme and the batch size - playing a crucial role in determining final performance.

Beyond the PCNs studied in this work, EP has applications in a broad range of models, including analog hardware systems where inference and gradient extraction are directly performed by the system's physics \citep{dillavou2022demonstration,yi2023activity,laydevant2024training}. Although it is unclear whether PCNs could be efficiently implemented onto such hardware - see~\citet{zahid2023predictive} for a critical review of this question - they may be useful for future developments of EP in at least two ways. First, PCNs allow us to evaluate the performance of EP on standard feedforward neural network models and compare with the BP baseline. Second, they can also be used as a baseline for physical systems of the same size: for instance, the construction by \citet{scellier2025universal} to convert an MLP into an approximately equivalent nonlinear resistor network provides a direct benchmark. Consequently, as we attempt to scale the training of physical systems with EP, PCNs constitute a useful concept to decouple EP-related challenges from model-specific issues. In this regard, our experimental results strengthen EP's standing as an effective training method, and suggest that current challenges encountered in scaling simulations of EP in physical networks might come more from the expressivity and trainability of these systems than from EP itself (or perhaps from the interplay between both).

\section*{Acknowledgements}

This work has been funded by Rain AI and the Advanced Research + Invention Agency's (ARIA) Scaling Compute programme.

\bibliography{biblio}
\bibliographystyle{plainnat}

\newpage
\appendix

\section{Nudge-Phase Dynamics: Traversal Scheme, Step Size, Modified PGD, and Equilibration Curves}
\label{sec:nudge-phase-equilibration}

In this appendix, we provide a complete description of the nudge-phase dynamics used in our experiments. In particular, we justify experimentally the choice $\alpha=1$ for the step size, and we present a modified version of PGD combined with an asynchronous traversal scheme, which we found to work better than PGD with synchronous updates. We also take a close look at the nudge-phase equilibration curves, i.e. the total energy ($F$) as a function of the number of nudge iterations ($K$).

\subsection{Traversal Scheme and Step Size}

Updating all the layers synchronously at each iteration of the PGD scheme does not work well in practice. Instead, we find that an `asynchronous update' scheme works much better: at every iteration, we first update the layers of even indices (one half of the layers) and then we update the layers of odd indices (the other half of the layers). Relaxing all the layers once (first the even layers, then the odd layers) constitutes one \textit{iteration}. We repeat as many iterations as is necessary until convergence to the nudged equilibrium. This asynchronous scheme is similar in spirit to the one used in Deep Boltzmann Machines \citep{salakhutdinov2009deep}, Deep Hopfield Networks \citep{scellier2023energy,goemaere2023accelerating} and Deep Resistive Networks \citep{scellier2024fast}.

\begin{figure}[h]
    \centering
    \includegraphics[width=1\linewidth]{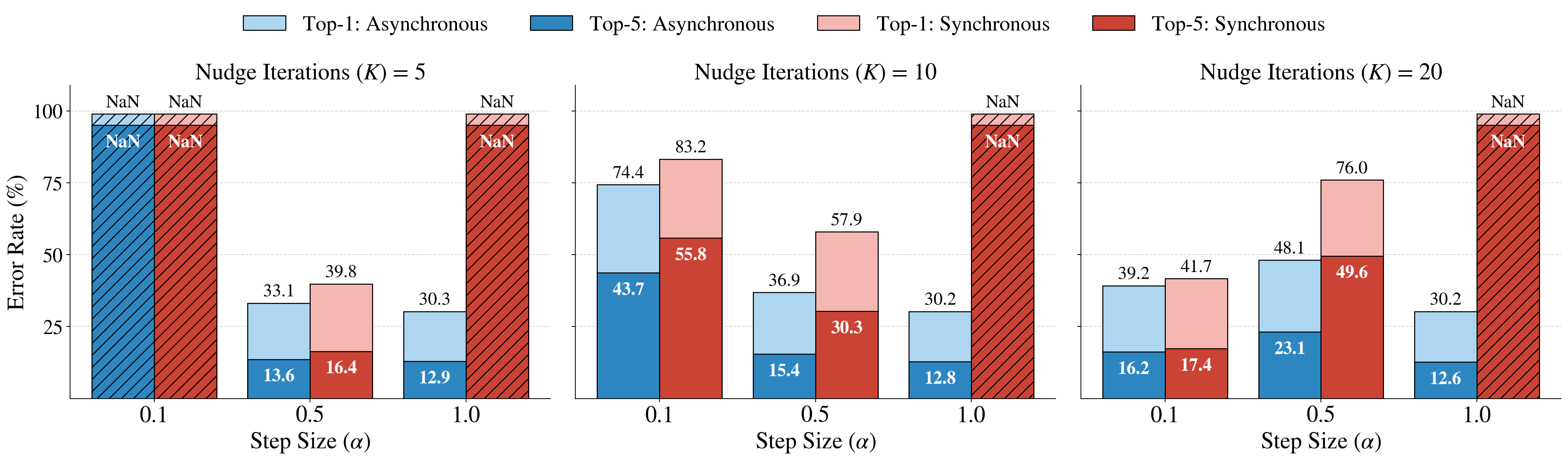}
    \caption{Test error rates obtained with the VGG5 trained on CIFAR100 using different traversal schedules (synchronous and asynchronous updates), number of nudged phase iterations $K$ (5, 10 and 20) and step sizes $\alpha$ (0.1, 0.5, 1.0). NaN means that training failed and diverged, resulting in NaN values.}
    \label{fig:step_size_and_traversal}
\end{figure}

Figure~\ref{fig:step_size_and_traversal} shows the performance (test error rates) of the VGG5 model trained on CIFAR-100 for different step sizes ($\alpha$), different number of nudge-phase iterations ($K$) and different traversal schemes (synchronous vs asynchronous). The asynchronous update scheme with step size $\alpha=1$ performs the best, and shows equivalent performance for 5, 10 and 20 nudged-phase iterations.

\subsection{Modified Projected Gradient Descent (mod-PGD)}
\label{sec:mod-pgd}

In our experiments, we used a modified version of PGD, which we found to work slightly better than PGD itself. Recall that the PGD update rule with step size $\alpha=1$ writes:
\begin{equation}
h_k \gets \text{ReLU}\left(\text{ReLU}(a_k)
+ \frac{\partial \epsilon_{k+1}^2}{\partial h_k}\right),
\label{eq:pgd2}
\end{equation}
where $a_k$ is a short hand for the pre-activations, defined such that ${\rm ReLU}(a_k) = f_k(\theta_k, h_{k-1})$. The `mod-PGD' update rule reads
\begin{align}
h_k \gets \text{ReLU}\left(a_k
+ \frac{\partial \epsilon_{k+1}^2}{\partial h_k}\right).
\label{eq:mod-pgd}
\end{align}
For the output layer, which does not use a nonlinearity, mod-PGD results in the same update rule as PGD.

\begin{figure}[htbp]
    \centering
    \begin{subfigure}[b]{0.3\textwidth}
        \centering
        \includegraphics[width=\textwidth]{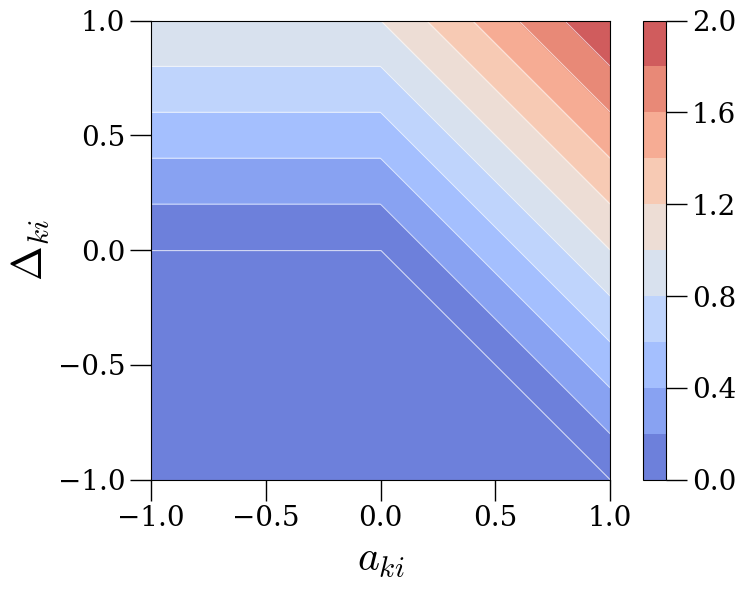}
        \caption{PGD}
        \label{fig:pgd-levels}
    \end{subfigure}
    \qquad\qquad
    \begin{subfigure}[b]{0.3\textwidth}
        \centering
        \includegraphics[width=\textwidth]{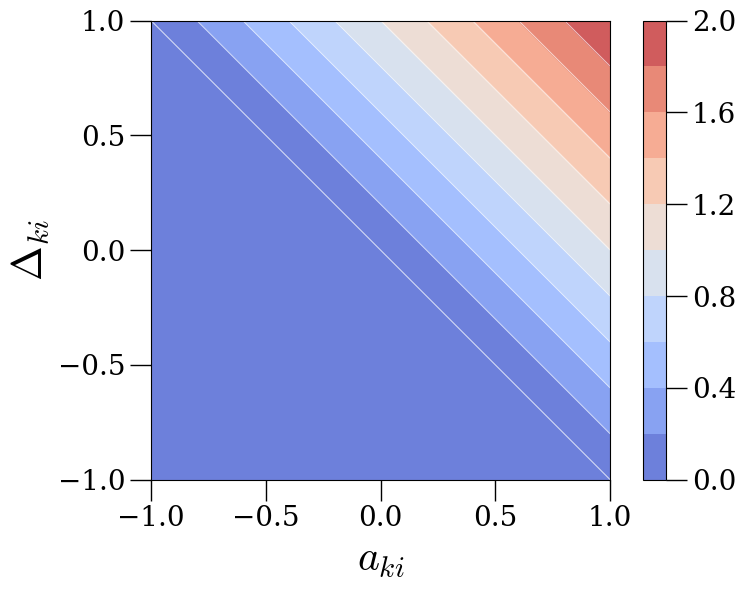}
        \caption{mod-PGD}
        \label{fig:mod-pgd-levels}
    \end{subfigure}
    \caption{Level curves of the update rules prescribed by PGD (RHS of Eq.~\eqref{eq:pgd2}) and mod-PGD (RHS of Eq.\eqref{eq:mod-pgd}). We consider a single unit $i$ in layer $k$.}
    \label{fig:level-curves}
\end{figure}

Figure~\ref{fig:level-curves} summarizes the behavior of the PGD~\eqref{eq:pgd2} and mod-PGD~\eqref{eq:mod-pgd} dynamics, as a function depending on the bottom-up signals to the the k-th layer ($a_k$) and the top-down signals to the k-th layer ($\Delta_k:=\tfrac{\partial \epsilon_{k+1}^2}{\partial h_k}$). For a unit $i$ in layer $k$, Equations~\eqref{eq:pgd2} and ~\eqref{eq:mod-pgd} only differ in the case where $a_{ki}<0$ and $\Delta_{ki} > 0$. This is reflected in Figure~\ref{fig:pgd-levels} and Figure~\ref{fig:mod-pgd-levels}, which only differ in the top left quadrant. Notably for positive $\Delta_{ki}$, the PGD dynamics ignores inhibitory bottom-up signals ($a_{ki}<0$). The mod-PGD dynamics on the other hand allows negative bottom-up signals to act as an inhibitory buffer, meaning bottom-up inhibited units remain inactive unless the top-down signal is sufficiently strong. By tempering the top-down signals in this manner, the mod-PGD dynamics remain closer to the free phase equilibrium than the PGD dynamics.

\begin{figure}[h]
    \centering
    \includegraphics[width=0.78\linewidth]{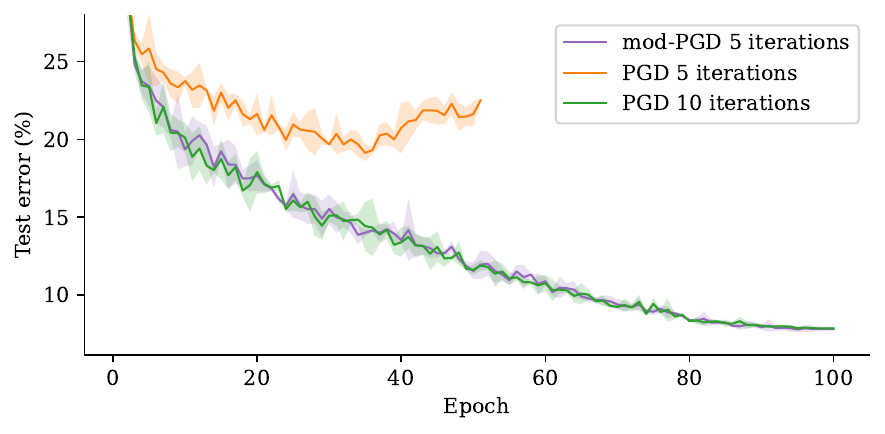}
    \caption{Comparison of PGD and mod-PGD on the VGG5 model trained on CIFAR10 using Nudge-MSE. Solid lines indicate the mean over three runs and shaded areas indicate $\pm$ one standard deviation. The PGD experiments with 5 iterations consistently diverged around epoch 50. These results correspond to those reported in Table~\ref{table:modified-projected-gradient-descent}.}
    \label{fig:PGD-vs_modPGD}
\end{figure}

Table~\ref{table:modified-projected-gradient-descent} compares PGD and mod-PGD experimentally. On CIFAR100 and ImageNet 32x32, mod-PGD leads to a very small, yet consistent improvement over PGD. On CIFAR10, however, PGD performs significantly worse than mod-PGD, or requires more nudge iterations (larger $K$). Figure~\ref{fig:PGD-vs_modPGD} shows the corresponding training curves on CIFAR-10. PGD with $K=10$ performs on par with mod-PGD with $K=5$, but PGD with $K=5$ terminates early because training eventually diverged resulting in NaN values.

\begin{table}[h]
\caption{Comparison of PGD and mod-PGD, where the VGG5 model is trained on different datasets using MSE-Nudge. The numbers in parenthesis indicate the number of iterations used in the nudge phase ($K$). Results were averaged over 3 random seeds.}
\label{table:modified-projected-gradient-descent}
\centering
\setlength{\tabcolsep}{3.5pt}
\begin{tabular}{cccccc}
\toprule
Algorithm & CIFAR-10 & \multicolumn{2}{c}{CIFAR-100} & \multicolumn{2}{c}{ImageNet 32x32} \\
(K iterations)& Top1 & Top1 & Top5 & Top1 & Top5 \\
\midrule
mod-PGD (K=5) & $7.82 \pm 0.14$ & $30.13 \pm 0.08$ & $12.80 \pm 0.11$ & $70.82 \pm 0.20$ & $55.93 \pm 0.28$\\
PGD (K=5) & $22.97 \pm 0.30$ & $30.47 \pm 0.09$ & $12.83 \pm 0.07$ & $71.38 \pm 0.19$ & $56.28 \pm 0.27$ \\
PGD (K=10) & $7.86 \pm 0.09$ & $30.20 \pm 0.27$ & $12.85 \pm 0.30$ & $71.30 \pm 0.11$ & $56.42 \pm 0.11$ \\
\bottomrule
\end{tabular}
\end{table}

\subsection{Nudge-Phase Equilibration Curves}

\begin{figure}[h]
    \centering
    \begin{subfigure}[b]{0.4\linewidth}
        \centering
        \includegraphics[width=\linewidth]{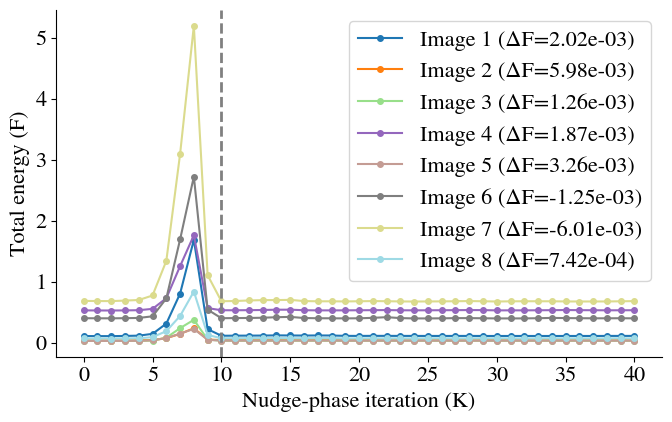}
        \caption{Mod-PGD with $\beta=0.05$}
        \label{fig:equilibration-mod-pgd-positive-beta}
    \end{subfigure}
    \begin{subfigure}[b]{0.4\linewidth}
        \centering
        \includegraphics[width=\linewidth]{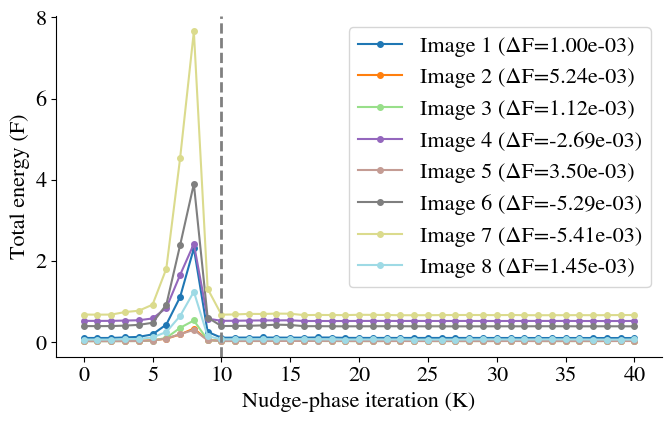}
        \caption{PGD with $\beta=0.05$}
        \label{fig:equilibration-pgd-positive-beta}
    \end{subfigure}
    \begin{subfigure}[b]{0.4\linewidth}
        \centering
        \includegraphics[width=\linewidth]{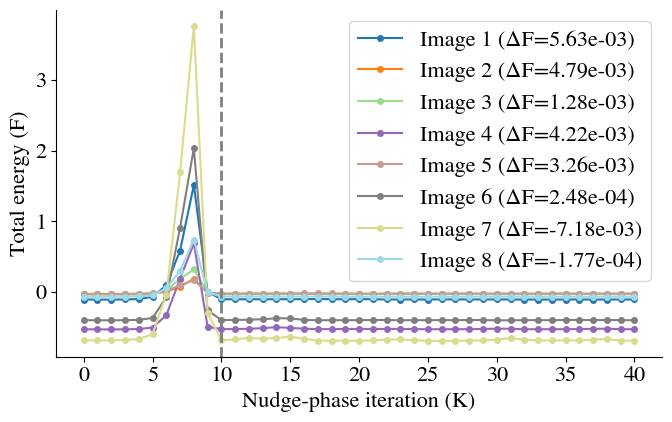}
        \caption{Mod-PGD with $\beta=-0.05$}
        \label{fig:equilibration-mod-pgd-negative-beta}
    \end{subfigure}
    \begin{subfigure}[b]{0.4\linewidth}
        \centering
        \includegraphics[width=\linewidth]{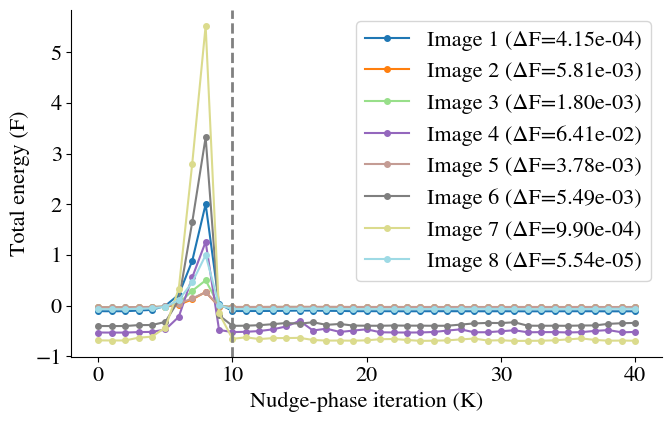}
        \caption{PGD with $\beta=-0.05$}
        \label{fig:equilibration-pgd-negative-beta}
    \end{subfigure}
    \caption{Nudge-phase equilibration curves for the VGG10 model evaluated on 8 random images from the ImageNet validation set. The plot represents the total energy $F$ as a function of the number of nudge iterations ($K$), for both mod-PGD and PGD as well as for both positive ($\beta=0.05$) and negative ($\beta=-0.05$) values of the nudging parameter. The dashed gray line corresponds to $K=10$, the value of $K$ used in our training experiments. The energy gap between the free state and nudge state is defined as $\Delta F = F(K=10) - F(K=0)$.}
    \label{fig:equilibration}
\end{figure}

Finally, we take a closer look at the nudge-phase equilibration dynamics. Figure~\ref{fig:equilibration} shows the total energy ($F = E + \beta C$) as a function of the number of nudge iterations ($K$) for the VGG10 model, using both PGD and mod-PGD. We make a few comments.

First, the equilibration curves show a significant spike in the total energy, starting around $K=5$ and stabilizing at $K=10$. These curves justify our choice of $K=10$ in the VGG10 experiments.

Second, using the asynchronous update scheme (where at each iteration one first updates the even layers and then the odd layers), the teaching signals introduced at the output layer at the beginning of the nudge phase propagate two layers per iteration. Hence, the beginning of the spike at $K=5$ coincides with the minimal number of iterations required for the teaching signals to propagate from the output layer to the input layer.

The spike is generally larger for PGD than for mod-PGD, which is consistent with our observation (Appendix~\ref{sec:mod-pgd}) that, in comparison with PGD, mod-PGD tempers top-down (error) signals.

Perhaps most importantly, the total energy does not decrease as a function of $K$ during the nudge-phase equilibration. In fact, recalling that the initial state of the network is the free equilibrium, the nudge equilibrium ($K=10$) often has higher energy than the free equilibrium ($K=0$), which translates as $\Delta F > 0$ in Table~\ref{fig:equilibration}. This shows that the nudge equilibrium is not, in general, the global minimum of $F$. EP, however, only requires the first-order stationary condition of Eq.~\eqref{eq:nudge-state}. Thus, the nudge equilibrium may as well be a local minimum or even a saddle point of $F$, and our PGD and mod-PGD dynamics might be better understood as fixed point iteration methods that find critical points of $F$, rather than as minimizing $F$. Similar observations were recently made in the Convergent Energy Transformer \citep{hoier2026training}.

We note that our nudge-phase equilibration approach is different from that of \citet{pinchetti2025benchmarking}, who employed GD with momentum, using tiny learning rates.

\clearpage
\section{Experimental Details}
\label{sec:simulation-details}

In this appendix, we provide the implementation details for our experiments.

\subsection{Datasets}

We use five datasets: MNIST, CIFAR-10, CIFAR-100, ImageNet32 and (full-size) ImageNet.

The MNIST dataset is composed of images of handwritten digits \citep{lecun1998gradient}. Each image $x$ in the dataset is a $28 \times 28$ gray-scaled image and comes with a label $y \in \left\{ 0, 1, \ldots, 9 \right\}$ indicating the digit that the image represents. The dataset contains 60,000 training images and 10,000 test images.

The CIFAR-10 dataset \citep{krizhevsky2009learning} consists of 60,000 colour images of $32 \times 32$ pixels. These images are split in 10 classes (each corresponding to an object or animal), with 6,000 images per class. The training set consists of 50,000 images and the test set of 10,000 images.

The CIFAR-100 dataset \citep{krizhevsky2009learning} also comprises 60,000 color images with a resolution of $32 \times 32$ pixels, featuring a diverse set of objects and animals. These images are categorized into 100 distinct classes, each containing 600 images. Like CIFAR-10, the dataset is divided into a training set with 50,000 images and a test set containing the remaining 10,000 images.

The ImageNet dataset \citep{deng2009imagenet} contains over a million high-resolution images, spanning thousands of object categories. Specifically, in the 2012 Large Scale Visual Recognition Challenge (ILSVRC2012), a popular subset of ImageNet, there are 1,000 distinct classes, each populated with a variable number of images. The ILSVRC2012 subset divides the dataset into a set for training, validation, and testing, facilitating comparisons across different approaches.

In addition to ImageNet (which here refers to the dataset with full-resolution images), we also consider the ImageNet32 variant, which consists of all images of the ImageNet dataset downsampled to 32x32 pixel images.

\paragraph{Data pre-processing and data augmentation.}

The data is normalized using the parameters provided in Table~\ref{table:data-normalization}.

\begin{table}[ht]
\caption{Data normalization. We normalize the input images using the recommended mean ($\mu$) and std ($\sigma$) values for each dataset. The MNIST images are gray-scale, i.e. they have a unique channel. The CIFAR-10, CIFAR-100 and ImageNet images are color images, i.e. they have three channels.}
\label{table:data-normalization}
\centering
\begin{tabular}{ccc}
\toprule
 & mean ($\mu$) & std ($\sigma$) \\
\cmidrule(r){1-3}
MNIST & 0.1307 & 0.3081 \\
CIFAR-10 & (0.4914, 0.4822, 0.4465) & (0.2023, 0.1994, 0.2010) \\
CIFAR-100 & (0.5071, 0.4867, 0.4408) & (0.2675, 0.2565, 0.2761) \\
ImageNet & (0.485, 0.456, 0.406) & (0.229, 0.224, 0.225) \\
\bottomrule
\end{tabular}
\end{table}

To make the VGG5 model compatible with MNIST, each image of the MNIST dataset (of size 28x28) is augmented to a 32x32-pixel image by adding two pixels at the top, two pixels at the bottom, two pixels to the left and two pixels to the right.

Finally, during training, we use random horizontal flipping and random cropping on CIFAR-10, CIFAR-100 and ImageNet. These stochastic augmentations are not used at test time. These stochastic augmentations are not used on MNIST either.

\subsection{VGG architectures}\label{app:vgg-architecture}

Table~\ref{table:vgg5} and Table~\ref{table:vgg10} contain the architectural details of the VGG5 and VGG10 models, respectively, as well as the hyperparameters used to obtain the results reported in Figure~\ref{fig:vgg5-batch-size}, Table~\ref{table:vgg5-results} and Table~\ref{table:vgg10-results}. Both VGG models use a combination of stacked convolutions and pooling layers followed by dense layers. All convolutional operations use stride 1, while max pooling operations use stride 2.

\begin{table}[ht]
    \centering
    
    \caption{VGG5 architecture. num\_inputs is 1 for MNIST, and 3 for other datasets (CIFAR10, CIFAR100 and ImageNet variants). num\_outputs is 10 for MNIST and CIFAR10, 100 for CIFAR100, and 1000 for ImageNet and ImageNet32.}
    \label{table:vgg5}
    \begin{tabular}{lccc}
        \toprule
        \textbf{Operation} & \textbf{Channels in} & \textbf{Channels out} & \textbf{Kernels} \\
        \midrule
        Conv2d & num\_inputs & $128$ & $3\times3$  \\
        \midrule
        Conv2d & $128$ & $256$ & $3\times3$   \\
        \midrule
        MaxPool & -- & -- & $2\times2$   \\
        \midrule
        Conv2d & $256$ & $512$ & $3\times3$   \\
        \midrule
        MaxPool & -- & -- & $2\times2$   \\
        \midrule
        Conv2d & $512$ & $512$ & $3\times3$   \\
        \midrule
        MaxPool & -- & -- & $2\times2$   \\
        \midrule
        Conv2d & $512$ & $512$ & $3\times3$   \\
        \midrule
        MaxPool & -- & -- & $2\times2$   \\
        \midrule
        Dense & \multicolumn{2}{c}{num\_outputs} \\
        \bottomrule
    \end{tabular}

    \vspace{1cm} 

    \caption{VGG10 architecture}
    \label{table:vgg10}
    \begin{tabular}{lccc}
        \toprule
        \textbf{Operation} & \textbf{Channels in} & \textbf{Channels out} & \textbf{Kernels} \\
        \midrule
        Conv2d & $3$ & $64$ & $3\times3$  \\
        \midrule
        MaxPool & -- & -- & $2\times2$   \\
        \midrule
        Conv2d & $64$ & $128$ & $3\times3$   \\
        \midrule
        MaxPool & -- & -- & $2\times2$   \\
        \midrule
        Conv2d & $128$ & $256$ & $3\times3$   \\
        \midrule
        Conv2d & $256$ & $256$ & $3\times3$   \\
        \midrule
        MaxPool & -- & -- & $2\times2$   \\
        \midrule
        Conv2d & $256$ & $512$ & $3\times3$   \\
        \midrule
        Conv2d & $512$ & $512$ & $3\times3$   \\
        \midrule
        MaxPool & -- & -- & $2\times2$   \\
        \midrule
        Conv2d & $512$ & $512$ & $3\times3$   \\
        \midrule
        Conv2d & $512$ & $512$ & $3\times3$   \\
        \midrule
        MaxPool & -- & -- & $2\times2$   \\
        \midrule
        Dense & \multicolumn{2}{c}{2048} \\
        \midrule
        Dense & \multicolumn{2}{c}{1000} \\
        \bottomrule
    \end{tabular}
\end{table}

\clearpage 

\paragraph{Weight initialization.} We initialize the weights of dense interactions and convolutional interactions according to
\begin{equation}
w_{ij} \sim \mathcal{U}(-c,+c), \qquad c = \gamma \sqrt{\frac{1}{\rm fan\_mode}}
\end{equation}
which is the `Kaiming uniform' scheme rescaled by a factor $\gamma$, that we call the `gain' here (i.e. a scaling number). For both VGG5 and VGG10, dense weights of shape (${\rm size}_{\rm pre}$, ${\rm size}_{\rm post}$), have ${\rm fan}_{\rm in} = {\rm size}_{\rm pre}$ ; for convolutional weights of shape (${\rm channel}_{\rm in}$, ${\rm channel}_{\rm out}$, ${\rm height}$, ${\rm width}$), VGG5 uses ${\rm fan}_{\rm in} = {\rm channel}_{\rm in} \times {\rm height} \times {\rm width}$ and VGG10 uses ${\rm fan}_{\rm out} = {\rm channel}_{\rm out} \times {\rm height} \times {\rm width}$

The gain $\gamma$ was consistently chosen to be $1$, except in the setting of the VGG5 trained on ImageNet32 using the MSE where we found that using $\gamma=0.2$ for the final (output) weight matrix led to significantly better results, as shown in Figure \ref{fig:vgg5-weight-init}.

\subsection{Hyperparameters related to training}

We train our VGG networks with both equilibrium propagation (EP) and backpropagation (BP). Training using BP is done the usual way. As for EP, at each training step, we proceed as follows. First we pick a mini-batch of samples in the training set, $x$, and their corresponding labels, $y$. Then we perform a forward pass to compute the free equilibrium, which we store. Assuming we use the centered scheme, let $\beta > 0$ be the nudging value used for training. First, we set the nudging parameter to $\beta$ and we perform $K$ asynchronous iterations of mod-PGD (Eq.~\eqref{eq:mod-pgd}). Next, we reset the state of the network to the free equilibrium, then we set the nudging parameter to $-\beta$, and we perform a new equilibration phase of $K$ asynchronous iterations of mod-PGD. Finally, we update all the parameters simultaneously in a single `parameter update' phase. See Table~\ref{table:vgg5hparams} and Table~\ref{table:vgg10hparams} for the hyperparameters used.

\paragraph{Optimizer and scheduler.}
We use mini-batch gradient descent (SGD) with momentum, weight decay and the Nesterov method. We also use a cosine-annealing scheduler for the learning rates with hyperparameters $T_{\rm max}$ and $\eta_{\rm min}$.

\paragraph{Hyperparameter Optimization.}
We used a two-stage hyperparameter tuning strategy, where we first tuned the hyperparameters shared by BP-based and EP-based training (weight initialization schemes, cost function, learning rates, weight decay, batch size, ...), and subsequently tuned the EP specific hyperparameters (perturbation approach, finite difference scheme, nudging sternght $\beta$, number of nudge iterations $K$, and PGD variant). This two-stage strategy is justified by the fact that, when employing the nudging-based perturbation approach, in the limit when $\beta$ tends to $0$, the EP gradients are equal to the BP gradients. Tuning was done manually with hyperparameter sweeps, but without Optuna or other black-box tools.

\paragraph{Computational resources.}
The code for the simulations uses Pytorch 2.9~\citep{Ansel_PyTorch_2_Faster_2024} and TorchVision 0.24~\citep{torchvision2016}.
The simulations were carried on Nvidia A100 GPUs. The VGG10 experiments on ImageNet using EP took 18 days (resp. 10 days) for the centered scheme (resp. backward scheme and random scheme) to complete on a single GPU. The BP experiment took 36 hours to complete. Each VGG5 experiment on ImageNet32 using EP took 17 hours to complete.

Our implementation is based on PyTorch's eager execution mode, and could likely be accelerated by using \emph{just in time} compilation (as is done in~\citep{pinchetti2025benchmarking}).


\begin{table}
\caption{Hyperparameter values used in the VGG5 experiments to obtain the results of Table~\ref{table:vgg5-results}.}
\label{table:vgg5hparams}
\centering
\begin{tabular}{ccccc}
\toprule
 & MNIST & CIFAR-10 & CIFAR-100 & ImageNet32 \\
\midrule
mini-batch size with MSE & 16 & 16 & 16 & 256 \\
mini-batch size cross-entropy (CE) & 64 & 64 & 64 & 256 \\
nudging ($\beta$) & \multicolumn{4}{c}{0.02}  \\
num. iterations of mod-PGD ($K$) & \multicolumn{4}{c}{5} \\
learning rate ($\eta$) with MSE & \multicolumn{4}{c}{0.04} \\
learning rate ($\eta$) with CE & \multicolumn{4}{c}{0.01} \\
momentum & \multicolumn{4}{c}{0.9} \\
weight decay & \multicolumn{4}{c}{3 e-4} \\
number of epochs & \multicolumn{4}{c}{100} \\
$T_{\rm max}$ (scheduler hyperparameter) & \multicolumn{4}{c}{100} \\
$\eta_{\rm min}$ (scheduler hyperparameter) & \multicolumn{4}{c}{1e-6} \\
\bottomrule
\end{tabular}
\end{table}

\begin{table}
\caption{Hyperparameters used for the VGG10 training experiments on ImageNet.}
\label{table:vgg10hparams}
\centering
\begin{tabular}{ccccc}
\toprule
 & ImageNet \\
\midrule
nudging strength ($\beta$) & 0.05 \\
number of nudge-phase iterations ($K$) & 10 \\
learning rates (for the trainable parameters) & 0.02 \\
momentum & 0.9 \\
weight decay & 2 e-4 \\
mini-batch size & 256 \\
number of epochs & 50 \\
cosine annealing scheduler parameter $T_{\rm max}$ & 50 \\
cosine annealing scheduler parameter $\eta_{\rm min}$ & 2.0 e-6 \\
\bottomrule
\end{tabular}
\end{table}

\subsection{Training Curves of the VGG10 Experiments on ImageNet}

\begin{figure}[hbt]
    \centering
    \begin{subfigure}[b]{0.49\textwidth}
        \centering\includegraphics[width=\textwidth]{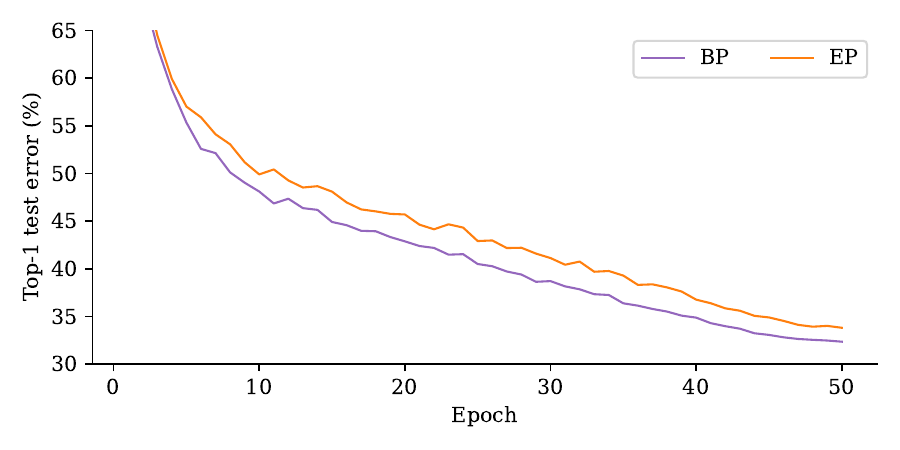}
        \caption{Top1 test error rate}
        \label{fig:vgg10-Top1}
    \end{subfigure}
    \hfill 
    \begin{subfigure}[b]{0.49\textwidth}
        \centering
        \includegraphics[width=\textwidth]{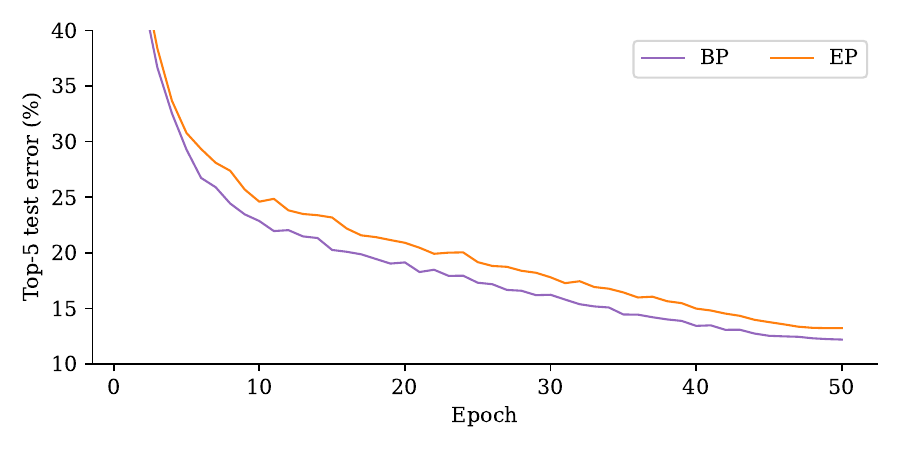}
        \caption{Top5 test error rate}
        \label{fig:vgg10-Top5}
    \end{subfigure}
    \caption{Training curves of the VGG10 experiments on full-size Imagenet. We observe a small gap between centered EP and BP.}
    \label{fig:vgg10-curves}
\end{figure}

Figure~\ref{fig:vgg10-curves} shows the training curves (Top1 and Top5 error rates) obtained with the VGG10 network trained on full-size ImageNet. 

\clearpage
\section{Traditional PCN Training, EP's Contrastive Function and the Penalty Method}
\label{sec:traditional-pcn-training}

In this appendix, we revisit several connections between PCNs, EP and the penalty method. In particular, we:
\begin{itemize}
\item derive the PCN energy function from its probabilistic interpretation,
\item review the traditional learning algorithm for PCNs,
\item relate the EP contrastive function to the standard PCN objective,
\item discuss the connection between EP-trained PCNs and the penalty method.
\end{itemize}

\subsection{Probabilistic Interpretation of PCNs}

The PCN energy function is commonly motivated from a probabilistic perspective. For each layer $k$, the state $h_k$ is assumed to be conditionally Gaussian given the previous layer:
\begin{equation}
h_k \sim \mathcal{N}\!\left(f_k(\theta_k,h_{k-1}), \sigma_k^2 I \right),
\end{equation}
where $f_k$ is the feedforward transformation parameterized by $\theta_k$, and $\sigma_k$ is the standard deviation. Equivalently, the conditional density of $h_k$ given $h_{k-1}$ is
\begin{equation}
P_{\theta_k}(h_k \mid h_{k-1}) = \frac{1}{(2\pi \sigma_k^2)^{d_k/2}} \exp\!\left( -\frac{1}{2\sigma_k^2} \left\| h_k - f_k(\theta_k,h_{k-1}) \right\|^2 \right),
\end{equation}
where $d_k$ denotes the dimensionality of layer $k$. The full network therefore defines the conditional distribution
\begin{equation}
P_\theta(h_1,\ldots,h_L \mid h_0=x) = \prod_{k=1}^L P_{\theta_k}(h_k \mid h_{k-1}),
\end{equation}
where $h_0=x$ is the network input. Given an input $x$, inference in a PCN consists of finding the most likely latent configuration:
\begin{equation}
\arg\max_h \log P_\theta(h \mid x),
\end{equation}
where $h=(h_1,\ldots,h_L)$ denotes the network state. Taking the negative log-likelihood yields
\begin{equation}
-\log P_\theta(h \mid x) = \sum_{k=1}^L \left[ \frac{d_k}{2}\log(2\pi\sigma_k^2) + \frac{1}{2\sigma_k^2} \left\| h_k - f_k(\theta_k,h_{k-1}) \right\|^2 \right].
\end{equation}
Choosing $\sigma_k=1$ for all $k$, the negative log-likelihood reduces, up to an additive constant, to the PCN energy function of Eq.~\eqref{eq:ff-compatible-energy-function}:
\begin{equation}
-\log P_\theta(h \mid x) = E_{\rm PCN}(\theta,x,h) + \mathrm{const}.
\end{equation}
Thus, maximizing the conditional likelihood is equivalent to minimizing the PCN energy function.

\subsection{Traditional Learning Algorithm for PCNs}

Under the probabilistic interpretation above, learning aims to maximize the conditional likelihood
\begin{equation}
\arg\max_\theta \log P_\theta(y \mid x),
\end{equation}
where $y$ is the desired target, associated with input $x$. However, evaluating this likelihood exactly requires marginalizing over all latent states:
\begin{equation}
\log P_\theta(y \mid x) = \log \int \exp\!\left( - E_{\rm PCN}(\theta,x,h_1,\ldots,h_{L-1},y) \right) dh_1 \cdots dh_{L-1},
\end{equation}
up to an additive constant. This integral is generally intractable. Traditional PCN learning therefore relies on a maximum a posteriori (MAP) approximation. Given a training pair $(x,y)$, the output layer is clamped to the target value $h_L=y$, and the latent states are inferred by minimizing the energy:
\begin{equation}
h_{1:L-1}^{\rm clamped} = \arg\min_{h_1,\ldots,h_{L-1}} E_{\rm PCN}(\theta,x,h_1,\ldots,h_{L-1},h_L=y).
\end{equation}
The MAP approximation assumes that the integral is dominated by the lowest-energy configuration:
\begin{equation}
\int e^{-E(h)}dh \approx e^{-E(h^{\rm clamped})}.
\end{equation}
Under this approximation,
\begin{equation}
-\log P_\theta(y \mid x) \approx E_{\rm PCN}(\theta,x,h_{1:L-1}^{\rm clamped},y) + \mathrm{const}.
\end{equation}
This leads to the objective
\begin{equation}
\mathcal{J}(\theta,x,y) = E_{\rm PCN}(\theta,x,h_{1:L-1}^{\rm clamped},y),
\end{equation}
which approximates the negative conditional log-likelihood up to an additive constant. The traditional PCN learning rule is obtained by differentiating this objective with respect to the parameters:
\begin{align}
\label{eq:standard-gradient-PCN}
\nabla_\theta \mathcal{J}(\theta,x,y) & = \frac{\partial E_{\rm PCN}}{\partial \theta} (\theta,x,h_{1:L-1}^{\rm clamped},y) + \frac{\partial E_{\rm PCN}}{\partial h} (\theta,x,h_{1:L-1}^{\rm clamped},y) \cdot \frac{\partial h_{1:L-1}^{\rm clamped}}{\partial \theta} \\
& = \frac{\partial E_{\rm PCN}}{\partial \theta} (\theta,x,h_{1:L-1}^{\rm clamped},y),
\end{align}
where we use the first order equilibrium condition $\frac{\partial E_{\rm PCN}}{\partial h} (\theta,x,h_{1:L-1}^{\rm clamped},y) = 0$. This recovers the standard predictive coding update rule: neural activities are first relaxed toward a minimum of the energy function, after which synaptic parameters are updated locally using prediction errors and presynaptic activities.

\subsection{EP's Contrastive Function in PCNs}

Unlike traditional PCN learning, EP does not require approximating an intractable posterior distribution and naturally accommodates arbitrary differentiable cost functions. EP introduces an explicit cost function $C$ and defines the objective
\begin{equation}
\mathcal{L}(\theta,x,y) = C(h_\star^0,y),
\end{equation}
where the free equilibrium state $h_\star^0$ satisfies
\begin{equation}
\frac{\partial E}{\partial h} \left( \theta,x,h_\star^0 \right) = 0.
\end{equation}
The original EP formula (Eq.~\eqref{eq:ep-formula}) states that
\begin{equation}
\nabla_\theta \mathcal{L}(\theta,x,y) = \frac{1}{\beta} \left[ \frac{\partial F}{\partial \theta}(\theta,\beta,h_\star^\beta) - \frac{\partial F}{\partial \theta}(\theta,0,h_\star^0) \right] + O(\beta),
\end{equation}
where the total energy and nudged equilibrium are defined as:
\begin{equation}
F(\theta,\beta,h) = E(\theta,x,h) + \beta C(h,y), \qquad \frac{\partial F}{\partial h} \left( \theta,\beta,h_\star^\beta \right) = 0.
\end{equation}
A more precise statement~\citep{scellier2023energy} shows that EP performs exact gradient descent on a contrastive objective,
\begin{equation}
\mathcal{L}_\beta(\theta,x,y) = \frac{1}{\beta} \left[ F(\theta,\beta,h_\star^\beta) - F(\theta,0,h_\star^0) \right],
\end{equation}
satisfying
\begin{gather}
\mathcal{L}_\beta(\theta,x,y) = \mathcal{L}(\theta,x,y) + O(\beta), \qquad \text{and} \\
\nabla_\theta \mathcal{L}_\beta(\theta,x,y) = \frac{1}{\beta} \left[ \frac{\partial F}{\partial \theta}(\theta,\beta,h_\star^\beta) - \frac{\partial F}{\partial \theta}(\theta,0,h_\star^0) \right].
\end{gather}

\begin{figure}[h]
\centering
\includegraphics[width=0.8\linewidth]{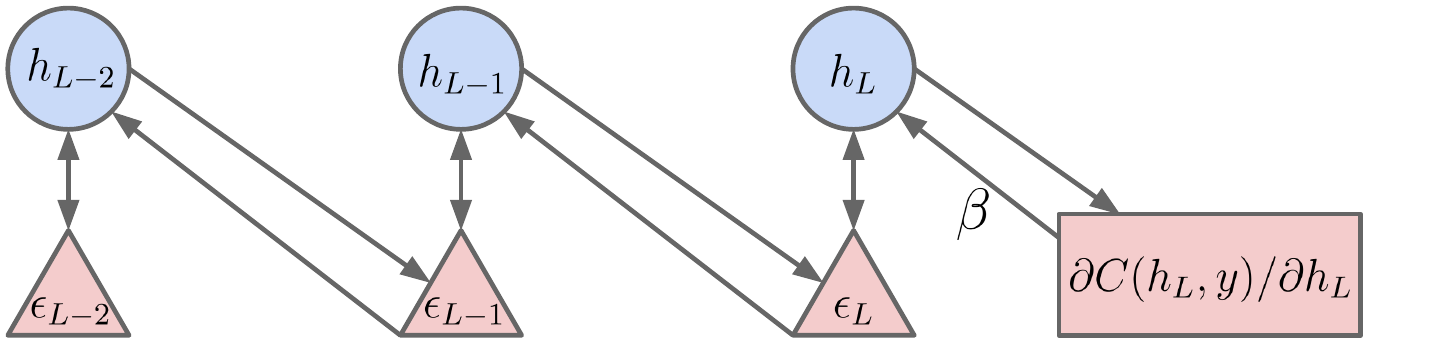}
\caption{Illustration of the nudging-based perturbation approach of EP in a PCN. In contrast to the clamping-based approach typically employed in PCNs, it allows to choose the cost function $C$ to optimize. Layers of units are represented by $h_k$, while layer-wise errors are denoted as $\epsilon_k = h_k - f_k(h_{k-1}, \theta_k)$. The total energy is the sum of squared errors and cost function, $F_{\rm PCN} = \frac{1}{2} \sum_{k=1}^L \left\| \epsilon_k \right\|^2 + C(h_L,y)$. During inference (free phase), the nudging parameter $\beta$ is set to zero, so that no teaching signal enters the network. During training (nudge phase), $\beta$ is set to a non-zero value, so that $-\beta \tfrac{\partial C}{\partial h_L}$ nudges the output units $h_L$ as in Eq.~\eqref{eq:output-units}. The centered scheme, which performs best in our experiments, requires two nudged equilibria: one with $\beta>0$ and one with $\beta<0$. The diagram is adapted from~\citet{millidge2022predictive}.}
\label{fig:pcn}
\end{figure}

For PCNs, the free equilibrium satisfies $E_{\rm PCN}(\theta,h_\star^0) = 0$, so that $F(\theta,0,h_\star^0) = 0$. The EP contrastive objective therefore simplifies to
\begin{equation}
\label{eq:contrastive-function}
\mathcal{L}_\beta(\theta,x,y) = \frac{1}{\beta} F(\theta,\beta,h_\star^\beta) = \min_h \left[ \frac{1}{\beta} \sum_{k=1}^L \left\| h_k - f_k(\theta_k,h_{k-1}) \right\|^2 + C(h_L) \right].
\end{equation}
This objective is illustrated in Figure~\ref{fig:pcn}. The EP gradient also simplifies:
\begin{gather}
\nabla_\theta \mathcal{L}_\beta(\theta,x,y) = \frac{1}{\beta} \frac{\partial F}{\partial \theta}(\theta,\beta,h_\star^\beta).
\end{gather}
To see this, we note that $\frac{\partial F}{\partial h}(\theta,0,h_\star^0) = 0$ by definition, since $h_\star^0$ is a stationary point. Moreover, $\frac{d}{d \theta}F(\theta,0,h_\star^0) = 0$ since $F(\theta,0,h_\star^0)$ is identically zero as a function of $\theta$. Therefore $\frac{\partial F}{\partial \theta}(\theta,0,h_\star^0) = \frac{d}{d \theta}F(\theta,0,h_\star^0) - \frac{\partial F}{\partial h}(\theta,0,h_\star^0) \cdot \frac{\partial h_\star^0}{\partial \theta} = 0$. 

As in Eq.~\eqref{eq:standard-gradient-PCN}, EP-based learning in PCNs proceeds in two stages: first, the network state is relaxed toward a stationary state of the total energy, and second, the parameters are updated using local derivatives evaluated at equilibrium.

\subsection{Relation to the Penalty Method}

The function
\begin{equation}
\frac{1}{\beta}F(\theta,\beta,h) = C(h_L,y) + \frac{1}{\beta} \sum_{k=1}^L \left\| h_k - f_k(\theta_k,h_{k-1}) \right\|^2
\end{equation}
appearing in the definition of $\mathcal{L}_\beta(\theta)$ can be viewed as a penalty function associated with the following constrained minimization problem:
\begin{equation}
\text{minimize } C(h_L), \qquad \text{subject to } h_k = f_k(\theta_k,h_{k-1}), \qquad 1 \leq k \leq L.
\end{equation}
This essentially recovers the penalty method. As the coefficient $\lambda = 1 / \beta$ tends to $+\infty$, the unconstrained problem is equivalent to the constrained problem.

Similar ideas to train feedforward networks using alternatives to backpropagation have been explored in multiple works \citep{carreira2014distributed,taylor2016training,gotmare2018decoupling,zach2021bilevel}.


\end{document}